\documentclass[5p,twocolumn,authoryear,nonatbib]{elsarticle}
\usepackage{amsmath,amssymb,booktabs,graphicx,placeins,array,xcolor,fancyvrb}
\newcommand{\citep}[1]{\csname papercitep@#1\endcsname}
\newcommand{\citet}[1]{\csname papercitet@#1\endcsname}
\expandafter\def\csname papercitep@llava15,qwen25vl,internvl35\endcsname{{(Bai et al., 2025; H. Liu et al., 2024; W. Wang et al., 2025)}}
\expandafter\def\csname papercitep@prompt_flaws\endcsname{{(Rosenthal et al., 2026)}}
\expandafter\def\csname papercitep@choices,length_bias,spotlight,stable_rag\endcsname{{(Atabuzzaman et al., 2026; Cho et al., 2026; Yao et al., 2025; Zhang et al., 2026)}}
\expandafter\def\csname papercitep@zong_permutations\endcsname{{(Zong et al., 2024)}}
\expandafter\def\csname papercitep@order_matters,multimodal_order\endcsname{{(Tan et al., 2024; Wei et al., 2026)}}
\expandafter\def\csname papercitep@modality_ttt\endcsname{{(Gupta \& Gandelsman, 2026)}}
\expandafter\def\csname papercitep@pbm_lora,teacher_student\endcsname{{(Guda et al., 2025; Liusie et al., 2024)}}
\expandafter\def\csname papercitep@positional_overload,set_llm\endcsname{{(Egressy \& Stühmer, 2025; Kinder et al., 2025)}}
\expandafter\def\csname papercitep@positional_overload\endcsname{{(Kinder et al., 2025)}}
\expandafter\def\csname papercitep@order_effect\endcsname{{(Guan et al., 2025)}}
\expandafter\def\csname papercitep@zong_permutations,lv_selection,revas\endcsname{{(Atabuzzaman et al., 2025; H. Wang et al., 2024; Zong et al., 2024)}}
\expandafter\def\csname papercitep@order_matters,multimodal_order,modality_ttt\endcsname{{(Gupta \& Gandelsman, 2026; Tan et al., 2024; Wei et al., 2026)}}
\expandafter\def\csname papercitet@zong_permutations\endcsname{{Zong et al. (2024)}}
\expandafter\def\csname papercitet@teacher_student\endcsname{{Liusie et al. (2024)}}
\expandafter\def\csname papercitep@zong_permutations,pbm_lora,gps\endcsname{{(Fatemi et al., 2026; Guda et al., 2025; Zong et al., 2024)}}
\expandafter\def\csname papercitep@pbm_lora,teacher_student,pa_grpo,dgao\endcsname{{(Chu et al., 2026; Guda et al., 2025; Liusie et al., 2024; Zheng et al., 2026)}}
\expandafter\def\csname papercitep@set_llm,invarirank,candidate_independent\endcsname{{(Amit, 2026; Bito et al., 2026; Egressy \& Stühmer, 2025)}}
\expandafter\def\csname papercitep@kd\endcsname{{(Hinton et al., 2015)}}
\expandafter\def\csname papercitep@lora\endcsname{{(Hu et al., 2022)}}
\expandafter\def\csname papercitep@teacher_student\endcsname{{(Liusie et al., 2024)}}
\expandafter\def\csname papercitep@two_factors,ceiling_bias\endcsname{{(Ok et al., 2026; Tamba, 2026)}}
\expandafter\def\csname papercitep@abcd,lv_selection,anchored\endcsname{{(Atabuzzaman et al., 2025; R. Li \& Gao, 2025; Nowak et al., 2026)}}
\expandafter\def\csname papercitep@rope\endcsname{{(Su et al., 2024)}}
\expandafter\def\csname papercitep@llava15\endcsname{{(H. Liu et al., 2024)}}
\expandafter\def\csname papercitep@qwen25vl\endcsname{{(Bai et al., 2025)}}
\expandafter\def\csname papercitep@internvl35\endcsname{{(W. Wang et al., 2025)}}
\expandafter\def\csname papercitep@vstar_data\endcsname{{(Wu \& Xie, 2024)}}
\expandafter\def\csname papercitep@aokvqa_data\endcsname{{(Schwenk et al., 2022)}}
\expandafter\def\csname papercitep@ocrvqa_data,future_attention\endcsname{{(Mishra et al., 2019; Pei et al., 2025)}}
\expandafter\def\csname papercitep@pope\endcsname{{(Y. Li et al., 2023)}}
\expandafter\def\csname papercitep@celebahq_data\endcsname{{(Karras et al., 2018)}}
\expandafter\def\csname papercitep@mmbench_data\endcsname{{(Y. Liu et al., 2024)}}
\expandafter\def\csname papercitep@seedbench\endcsname{{(B. Li et al., 2024)}}

\usepackage{xurl}
\usepackage{hyperref}
\hypersetup{colorlinks=true,linkcolor=blue,citecolor=blue,urlcolor=blue,pdfauthor={Yifei Cheng, Xiaolin Zhang, Yuxuan Qiu},pdftitle={Reducing Permutation Sensitivity in Multimodal Large Language Models via Shared Positions and Attention-Only Distillation}}

\makeatletter
\let\ps@pprintTitle\ps@plain
\g@addto@macro\elsaddress{\upshape}
\makeatother

\begin{document}

\begin{frontmatter}

\title{\textbf{Reducing Permutation Sensitivity in Multimodal Large Language Models via Shared Positions and Attention-Only Distillation}}

\author[bit]{Yifei Cheng}
\author[sdust]{Xiaolin Zhang}
\author[bit]{Yu-Xuan Qiu\corref{corresponding}}
\cortext[corresponding]{Corresponding author.}
\ead{qiu@bit.edu.cn}
\address[bit]{Beijing Institute of Technology, Zhuhai, China}
\address[sdust]{Shandong University of Science and Technology, China}

\begin{abstract}
In this paper, we conduct a systematic study on improving permutation robustness while preserving accuracy. First, we categorize permutation bias phenomena into four representative scenarios. Then, we take the multiple-choice question answering (MCQA) dataset as an example under text-option reordering scenario to analyze the root cause of permutation bias. Based on this root-cause analysis and set encoding strategy, we investigate shared positions with blocked cross-unit attention (SP-B) and extend this design to bidirectional cross-unit attention (SP-V) for modality exchange. However, these structural changes may reduce predictive accuracy. To recover model capability across different scenarios under the SP structure, an attention-only distillation from frozen native models is proposed. Shared positions reduce permutation sensitivity, and attention-only distillation improves accuracy under the shared-position structure. Across all three models, our method is comparable to or better than majority voting in macro-average all-permutation accuracy, and outperforms supervised fine-tuning by 1.95–4.20 percentage points. This work provides a valuable insight into permutation bias mitigation and model capability recovery.
\end{abstract}

\begin{keyword}Multimodal large language models \sep Permutation bias \sep Shared positional encoding \sep Knowledge distillation\end{keyword}

\end{frontmatter}

\section{Introduction}

Multimodal large language models (MLLMs) \citep{llava15,qwen25vl,internvl35} perform visual perception and reasoning within a unified language interface. Their capabilities motivate applications in document understanding, visual inspection, and intelligent decision support \citep{prompt_flaws}. Such systems must serialize images, questions, candidate answers, and retrieved examples before inference. Reliability therefore depends not only on recognizing task-relevant content, but also on avoiding sensitivity to incidental choices in how that content is presented \citep{choices,length_bias,spotlight,stable_rag}.

Despite their remarkable capabilities, MLLMs can produce inconsistent predictions when semantically equivalent inputs are presented in different orders, a phenomenon known as permutation bias \citep{zong_permutations}. Beyond candidate-answer reordering, similar sensitivities have been observed in the ordering of multimodal contexts and image--text demonstrations \citep{order_matters,multimodal_order}. Recent research further indicates that exchanging the relative positions of images and textual questions can affect model performance, even when the underlying task remains unchanged \citep{modality_ttt}. Such permutation sensitivity suggests that MLLMs may depend on the order in which the input is presented, rather than only on its task-relevant semantic meaning, raising concerns about their robustness and reliability in real-world applications.

Few studies have analyzed the root cause of permutation bias in MLLMs. Existing mitigation strategies are mainly designed for text-based LLMs and include inference-time permutation aggregation  \citep{zong_permutations}, training-based adaptation  \citep{pbm_lora,teacher_student}, and structural interventions \citep{positional_overload,set_llm}. These approaches leave a key question unresolved: how can an MLLM deliver consistently correct predictions across equivalent input orders while preserving pretrained capabilities and avoiding permutation aggregation? Addressing this question requires understanding the root cause and connecting the positional and attention relationships among multimodal units with the model's ability to use those units for prediction.

\begin{figure*}[tp]\centering\includegraphics[width=\textwidth,height=0.82\textheight,keepaspectratio]{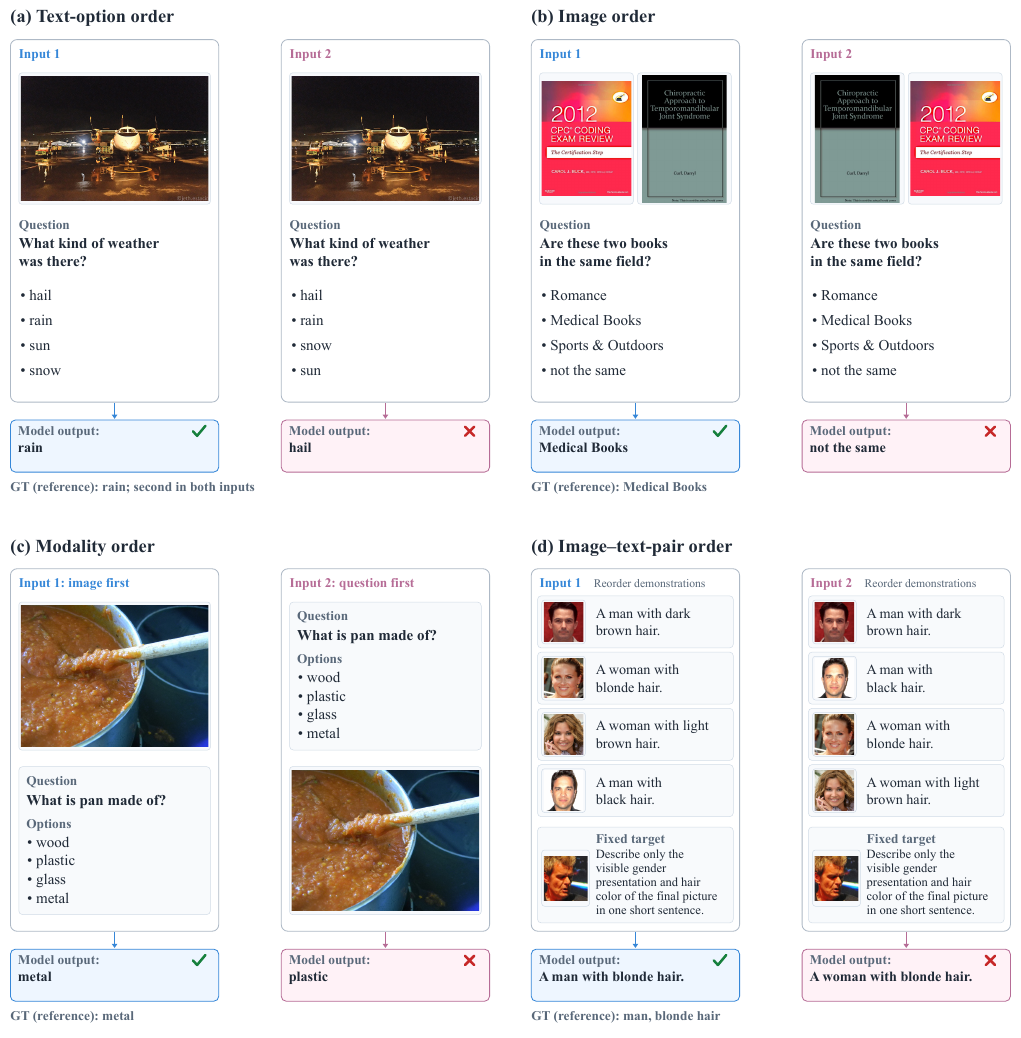}\caption{Illustration of four permutation sensitivity scenarios: (a) text-option reordering; (b) image reordering; (c) modality exchange; (d) image--text demonstration reordering.}\label{fig:tasks}
\par\smallskip\begin{minipage}{0.97\textwidth}\small LLaVA outputs are shown. Only the order of the marked units changes; the question and target answer stay fixed.\end{minipage}
\end{figure*}

Permutation sensitivity in MLLMs can be broadly categorized into four representative scenarios, as illustrated in Figure~\ref{fig:tasks}: (1) text-option reordering \citep{zong_permutations}, (2) image reordering, (3) modality exchange \citep{order_matters,multimodal_order}, and (4) image--text demonstration reordering \citep{order_matters,multimodal_order}. Specifically, Figure~\ref{fig:tasks}(a) shows that rearranging candidate answers changes the model's prediction from the correct answer ``rain'' to ``hail'', even though ``rain'' remains in the same position. In Figure~\ref{fig:tasks}(b), exchanging two book images changes the prediction from ``Medical Books'' to ``not the same''. Similarly, Figure~\ref{fig:tasks}(c) demonstrates that swapping the image and question blocks changes the prediction from the correct answer ``metal'' to ``plastic'', while Figure~\ref{fig:tasks}(d) shows that reordering image--text demonstrations changes the generated description from ``A man with blonde hair'' to ``A woman with blonde hair''. Across all four scenarios, the input content, task semantics, and ground-truth answers remain unchanged, yet the models produce inconsistent predictions solely due to input reordering.

In this work, our research can be summarized as follows. First, to investigate the root cause of permutation bias, we summarize four typical manifestations in Figure 1, formulate the corresponding scenarios in Section 3.1, and define a series of metrics in Section 3.2 to evaluate model performance comprehensively. We then use multiple-choice question answering (MCQA) as an example to study root cause in permutation bias. In Section 3.3, we experimentally demonstrate that permutation bias exists in both conventional labeled MCQA and option-label-free MCQA. Based on option-label-free MCQA, we construct a diagnostic subset in Section 3.4 to investigate the root cause of permutation bias in the subsequent section. Sections 4.1 and 4.3 present a mechanistic analysis of permutation sensitivity across three different models. The results show that positional encoding, causal masking, and numerical precision contribute to the observed bias. Second, based on this root-cause analysis and Set Encoding for LLMs (Kinder et al., 2025), we adapt shared positions with blocked cross-unit attention (SP-B) to multimodal inputs in Section 4.2. We then extend SP-B to bidirectional cross-unit attention (SP-V) in Section 4.4 to allow direct interaction between images and questions. Third, we describe experiments across the four typical scenarios in Section 5.1 to evaluate the shared-position structure. The results in Section 5.2 show that SP-B achieves strong permutation invariance, while those in Section 5.3 show that SP-V improves accuracy over SP-B in most evaluated modality-exchange settings. Both sections also report decreases in average accuracy. In Section 5.4, we design a teacher–student knowledge distillation framework to help models recover their capabilities under the shared-position structure and evaluate it in Section 5.5. An ablation comparison in Section 5.6 examines the effect of each module. Finally, Section 5.7 presents a comprehensive comparison of four methods: the baseline, training-free majority voting (Zong et al., 2024), shared positions with knowledge distillation, and shared positions with fine-tuning (Kinder et al., 2025). This comparison shows higher macro-average AA and AC than supervised fine-tuning across all three models, together with single-order inference.

The main contributions of this work are summarized as follows:

\noindent\textbf{1.}We categorize permutation bias into four representative scenarios (text-option, image, modality, and image–text demonstration reordering) and establish a unified evaluation framework with a series of multimodal metrics.

\noindent\textbf{2.}We conduct a root-cause study of permutation bias using text-order reordering in MCQA as a case study, and identify positional encoding, causal masking, and numerical precision as the contributing factors, with the first two being dominant.

\noindent\textbf{3.}We adapt the Set Encoding design to multimodal inputs as SP-B and extend it to SP-V to allow direct interaction between exchangeable units. While SP-B achieves strong permutation invariance, SP-V improves accuracy in modality-exchange settings.

\noindent\textbf{4.}We propose an attention-only knowledge distillation framework to recover model capability under the shared-position structure. Across three models, attention-only distillation achieves higher macro-average AA and AC than supervised fine-tuning \citep{positional_overload} using the same training and validation sets. It also exceeds majority voting \citep{zong_permutations} on both metrics for Qwen and InternVL.

\section{Related work}

\subsection{Permutation Sensitivity in LLMs and MLLMs}

Permutation sensitivity has been widely observed in both large language models (LLMs) and multimodal large language models (MLLMs) \citep{order_effect}. Previous studies have shown that rearranging answer options can substantially affect model predictions, revealing positional preferences and inconsistencies in multiple-choice reasoning \citep{zong_permutations,lv_selection,revas}. Beyond text-based inputs, recent studies have identified similar sensitivities in multimodal contexts, including image--text demonstration ordering, modality placement, and multimodal reasoning tasks \citep{order_matters,multimodal_order,modality_ttt}. These findings indicate that permutation sensitivity extends beyond text-only reasoning and represents a broader reliability challenge for MLLMs.

Prior work evaluates permutation bias through predictive accuracy, consistency across input orders, and systematic preferences for particular answer positions. \citet{zong_permutations} assess robustness using adversarial permutations, counting a question as correct only when every evaluated ordering yields the correct answer. \citet{teacher_student} distinguish positional bias from permutation sensitivity and quantify the latter using distances between content-aligned answer distributions. Building on these perspectives, we jointly report Original Accuracy (OA), Average Accuracy across permutations (AA), All-permutation Correct Accuracy (AC), and Invariant Predictions (IP). These complementary measures distinguish original-order performance, average correctness, correctness sustained across permutations, and prediction consistency irrespective of correctness. For mechanistic diagnosis, we additionally use Content L1, the mean L1 distance between content-aligned candidate distributions and their permutation average, to measure distributional changes that may occur without changing the selected answer. More detailed definitions of these metrics are given in Section 3.2.

\subsection{Methods for Reducing Permutation Bias}

Existing methods reduce permutation bias through inference-time aggregation, training, or changes to model structure. Inference-time methods combine predictions from different input orders or adjust answer scores \citep{zong_permutations,pbm_lora,gps}. Majority voting (MV) is a common example. It requires no parameter updates but evaluates multiple orders for each question. We use MV as an inference-time comparison.

Training-based methods use fine-tuning, teacher--student learning, or permutation-aware objectives to reduce order sensitivity \citep{pbm_lora,teacher_student,pa_grpo,dgao}. These studies show that training can improve consistency across input orders. In our method, shared positions and attention masks provide the structural basis for consistency. Training then helps the model recover accuracy under this structure.

Structural methods change positional encoding or attention visibility. Positional Overload introduces Set Encoding, which assigns shared positions to set elements and blocks attention between different elements \citep{positional_overload}. Related work also studies set representations and candidate isolation \citep{set_llm,invarirank,candidate_independent}. We adapt Set Encoding to multimodal inputs as SP-B. We then extend it to SP-V, which allows bidirectional attention between exchangeable units. This extension supports direct interaction between images and questions. These changes can reduce order sensitivity, but they can also affect predictive accuracy. This motivates the performance recovery methods discussed next.

\subsection{Recovering Performance under Modified Structures}

Modifying positions and attention masks disrupts the input processing of a pretrained model. The model may therefore need training to adapt to the new structure. Set-encoding methods use supervised fine-tuning for this purpose \citep{positional_overload,set_llm}. We use an FT protocol adapted from Positional Overload as a direct comparison. FT and our method use the same training and validation question lists and the corresponding shared-position structures.

Our training method builds on knowledge distillation (KD) and low-rank adaptation (LoRA). KD transfers information through a teacher's output distribution \citep{kd}. LoRA updates a small set of added parameters while keeping the original weights frozen \citep{lora}. We use these established methods to adapt the attention projections of a shared-position student to the candidate preferences of its native model.

Teacher--student training has also been used to reduce permutation bias in LLMs \citep{teacher_student}. Our KD objective focuses on recovering predictive accuracy after the structural change. The frozen native model provides one candidate distribution per training question under the original input order. The student uses SP-B or SP-V and learns from this distribution through a KL loss. Only attention LoRA parameters are updated. Unlike supervised FT, the KD loss uses teacher distributions rather than ground-truth answers. Training uses original-order MCQA and does not directly optimize agreement across permutations. Thus, the shared-position structure reduces order dependence, while KD helps the model retain its predictive ability.

\section{Problem Formulation and Evaluation Protocol}\label{sec:formulation}

Prior work reports that smaller models within the same family are generally more vulnerable to adversarial answer permutations \citep{zong_permutations}. Hence, we use three compact open-weight MLLMs—LLaVA-1.5-7B \citep{llava15}, Qwen2.5-VL-3B-Instruct \citep{qwen25vl}, and InternVL3.5-4B-HF \citep{internvl35}—for their relatively small parameter volume and accessible weights, which allow direct interventions on the inference structure.

In this section, we formally define the permutation scenarios (Section 3.1) and present the evaluation metrics used throughout this paper (Section 3.2). We also introduce an option-label-free protocol for MCQA (Section 3.3) and construct a diagnostic subset for mechanistic analysis (Section 3.4).

\subsection{Problem Formulation}\label{sec:units}

Write an input as $X=(P,E_1,\ldots,E_n,S)$, with fixed prefix $P$, exchangeable complete units $E_i$, and fixed suffix $S$. We test $X_\pi=(P,E_{\pi(1)},\ldots,E_{\pi(n)},S)$ only when the expected answer is unchanged. Word order within a unit, image geometry, and image--text associations are preserved.

MCQA can use explicit option letters with letter selection or unlabeled candidate rows with direct content selection. We refer to the latter as option-label-free MCQA. Section~\ref{sec:label-protocols} describes the detailed definition. Subsequent mechanism experiments and MCQA evaluations use option-label-free prompts.

\subsection{Evaluation Metrics}\label{sec:metrics}

To evaluate permutation sensitivity, we consider both prediction correctness and consistency across input permutations \citep{two_factors,ceiling_bias}. For question $i$, let $\Pi_i$ denote its valid permutation set, with $M_i=|\Pi_i|$. For each $\pi\in\Pi_i$, let $\hat y_{i\pi}$ denote the prediction and $a_{i\pi}\in\{0,1\}$ indicate correctness. Given $N$ questions, we define four metrics:

\begin{equation}\mathrm{OA}=\frac{1}{N}\sum_{i=1}^{N}a_{i\pi_0}.\label{eq:oa}\end{equation}

\begin{equation}\mathrm{AA}=\frac{1}{N}\sum_{i=1}^{N}\frac{1}{M_i}\sum_{\pi\in\Pi_i}a_{i\pi}.\label{eq:aa}\end{equation}

\begin{equation}\mathrm{AC}=\frac{1}{N}\sum_{i=1}^{N}\mathbb{I}\left[\forall\pi\in\Pi_i,\ a_{i\pi}=1\right].\label{eq:ac}\end{equation}

\begin{equation}\mathrm{IP}=\frac{1}{N}\sum_{i=1}^{N}\mathbb{I}\left[\left|\{\hat y_{i\pi}:\pi\in\Pi_i\}\right|=1\right].\label{eq:ip}\end{equation}

For each question, $\pi_0\in\Pi_i$ denotes its original input order. The function $\mathbb{I}[\cdot]$ is the indicator function. Original Accuracy (OA) measures performance under the original input order, while Average Accuracy (AA) evaluates the mean correctness across all permutations, assigning equal weight to each question. All-permutation Correct Accuracy (AC) measures the proportion of questions answered correctly under every permutation, reflecting robustness in terms of correctness. Invariant Predictions (IP) measures the proportion of questions for which the model produces identical predictions across all permutations, regardless of correctness.

Candidate predictions are compared by answer content, with fixed source identities for distinct candidates. Identical candidate strings are treated as the same semantic answer. CelebA-HQ IP compares complete decoded strings. By definition, AC cannot exceed OA or AA.

For controlled mechanistic analysis, we additionally use Content L1 to quantify variation in candidate probability distributions. Let $z_{i\pi}$ denote content-aligned candidate logits, $p_{i\pi}=\operatorname{softmax}(z_{i\pi})$, and $\bar{p}_i=M_i^{-1}\sum_{\pi\in\Pi_i}p_{i\pi}$. Content L1 is

\begin{equation}\mathrm{Content\ L1}=\frac{1}{N}\sum_{i=1}^{N}\frac{1}{M_i}\sum_{\pi\in\Pi_i}\left\|p_{i\pi}-\bar{p}_i\right\|_1.\label{eq:l1}\end{equation}

Unlike IP, which captures only whether the final predictions change, Content L1 measures the magnitude of variation in candidate probability distributions. This metric is applied exclusively to the restricted single-token diagnostic subset described in Section~\ref{sec:subset}.

\subsection{Removing Option Labels in MCQA}\label{sec:label-protocols}

Conventional multiple-choice question answering (MCQA) typically assigns letter labels (e.g., A/B/C/D) to candidate answers and requires the model to select the corresponding letter \citep{prompt_flaws}. Permuting candidate contents changes their associations with fixed option labels. These labels can introduce response preferences beyond candidate semantics \citep{abcd,lv_selection,anchored}. To reduce this confounding factor, we investigate an option-label-free protocol that removes explicit letter identifiers while preserving the original question, candidate contents, and ground-truth answer.

\begin{figure*}[tp]
\centering\includegraphics[width=\textwidth]{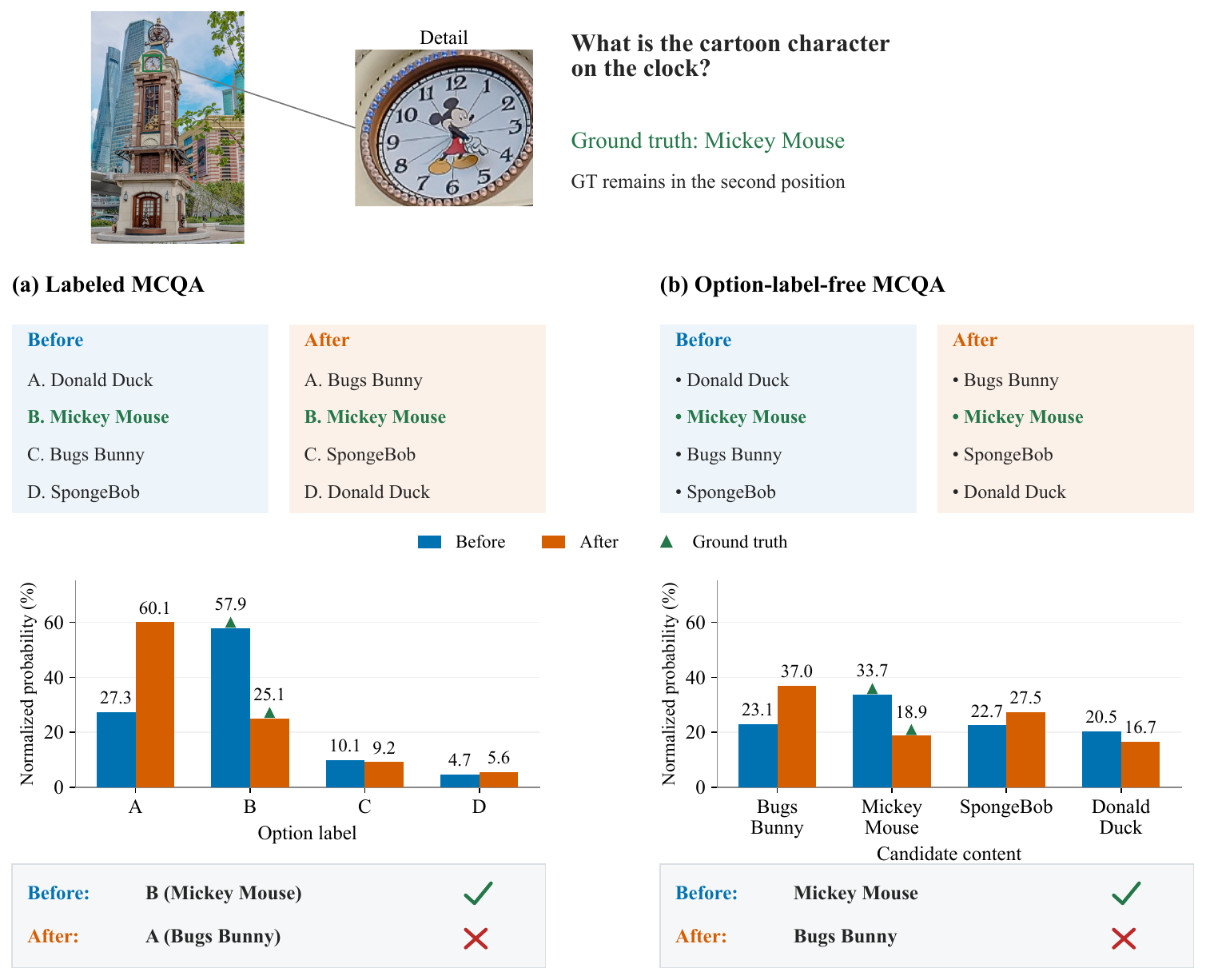}
\caption{Permutation sensitivity in (a) labeled and (b) option-label-free MCQA on the same V* question.}\label{fig:option-example}
\par\smallskip\begin{minipage}{0.97\textwidth}\small LLaVA scores option letters in (a) and candidate contents in (b). Bars show softmax-normalized mean token log-probabilities over the four possible answers. Both predictions change after reordering, even though the correct content stays in the same position.\end{minipage}

\end{figure*}

Figure~\ref{fig:option-example} shows permutation sensitivity while the correct answer stays in the same position. The cartoon character on the clock is Mickey Mouse, which remains the second candidate in both orders and retains label B in the labeled protocol. Only the three distractors are reordered. In both the labeled and option-label-free protocols, the prediction changes from the correct answer ``Mickey Mouse'' to the incorrect answer ``Bugs Bunny''. This change cannot be explained solely by moving the correct answer to a different position.

To evaluate the influence of option labels and response formats, we compare three protocols using LLaVA-1.5-7B on 191 V* questions: (1) letter-labeled inputs with letter scoring, (2) letter-labeled inputs with candidate-content scoring, and (3) option-label-free inputs with candidate-content scoring. All three protocols are evaluated under native positional encoding and causal attention in BF16 precision, covering the same 2,736 question--permutation pairs. Candidate scores are computed using the mean conditional token log-probability without appending EOS. For letter scoring, predictions are mapped back to their corresponding candidate contents before evaluation. For visualization, Figure~\ref{fig:option-example} expresses softmax-normalized answer scores as percentages.

\begin{table*}[t]
\centering\normalsize\renewcommand{\arraystretch}{1.18}
\caption{Comparison of MCQA input and scoring protocols.}\label{tab:label-protocols}
\setlength{\tabcolsep}{4pt}
\begin{tabular}{llrrrr}\toprule
Input & Scored output & OA $\uparrow$ & AA $\uparrow$ & AC $\uparrow$ & IP $\uparrow$\\\midrule
Labeled & Letter & 41.36 & 40.27 & 7.33 & 17.28\\
Labeled & Content & 47.64 & 44.70 & 23.04 & 46.07\\
Option-label-free & Content & \textbf{50.26} & \textbf{46.62} & \textbf{28.27} & \textbf{61.26}\\
\bottomrule\end{tabular}

\par\smallskip\begin{minipage}{0.97\textwidth}\small
LLaVA is evaluated on all 191 V* questions and all valid candidate orders. Values are percentages. Letter predictions are mapped back to answer content before scoring. OA: original-order accuracy; AA: mean accuracy across permutations; AC: accuracy requiring every order to be correct; IP: the share of questions with the same prediction across all orders. Bold marks the highest value in each column.
\end{minipage}
\end{table*}

As shown in Table~\ref{tab:label-protocols}, the option-label-free protocol achieves the highest AA (46.62\%), AC (28.27\%), and IP (61.26\%) among the three configurations. Direct content scoring improves consistency over letter-based selection, and removing explicit labels provides further improvements. Order sensitivity persists after removing option labels, establishing that the phenomenon extends beyond letter--content associations. Qwen and InternVL exhibit the same qualitative pattern as LLaVA: permutation bias occurs in both labeled and option-label-free MCQA (Supplementary Material, Section S2.1).

A labeled MCQA task can be answered through content selection followed by a deterministic mapping to the corresponding letter in the displayed ordering. Permutation-consistent content selection therefore supports semantically consistent answers in either output format. We consequently adopt option-label-free MCQA to study content-level order dependence without explicit option-letter associations.

\subsection{Diagnostic Subset for Mechanistic Analysis}\label{sec:subset}

To facilitate a controlled investigation of the mechanisms underlying permutation sensitivity, we construct a diagnostic subset from the V* benchmark. The original evaluation set contains 191 questions, including 107 four-choice and 84 two-choice questions. For LLaVA, we retain 75 four-choice questions whose candidate answers correspond to four distinct single-token IDs, excluding 32 four-choice questions that do not satisfy this criterion. Each retained question is evaluated under all 24 possible candidate permutations, yielding 1,800 question--permutation pairs per experimental condition for LLaVA. Applying the corresponding tokenization criterion to Qwen2.5-VL and InternVL3.5 yields 93 questions and 2,232 question--permutation pairs per condition for each model.

The single-token restriction enables direct alignment and comparison of candidate logits across permutations, avoiding additional complications introduced by multi-token answer scoring. Combined with the option-label-free protocol established in Section~\ref{sec:label-protocols}, this subset supports layer-wise analysis of candidate probability distributions and controlled interventions on positional encoding and attention visibility. The single-token criterion applies to output candidates. Input units include their formatting tokens. This subset is used exclusively for mechanistic diagnosis in Section~\ref{sec:mechanism}, while the broader evaluations in Section~\ref{sec:experiments} retain their original candidate configurations without imposing the single-token restriction.

\section{Mechanistic Analysis and Structural Mitigation of Permutation Sensitivity}\label{sec:mechanism}

In this section, we analyze the root cause of permutation sensitivity and locate its origin at the attention output (Section 4.1). Motivated by this finding, we propose the SP-B structure (Section 4.2) and investigate the coupled effects of positions and attention visibility (Section 4.3). Finally, we extend this design to SP-V for bidirectional modality exchange (Section 4.4).

\begin{figure*}[tp]\centering\includegraphics[width=\textwidth,height=0.40\textheight,keepaspectratio]{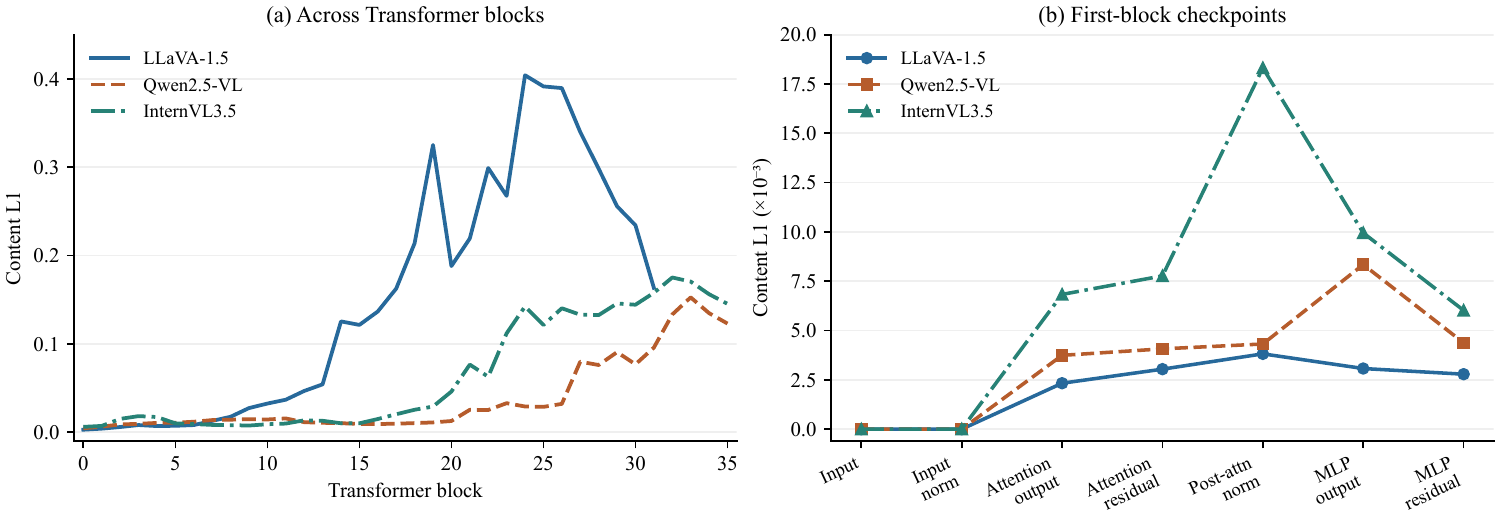}\caption{Layer-wise permutation sensitivity across three models: (a) block-end readouts; (b) first-block checkpoints.}\label{fig:layers}
\par\smallskip\begin{minipage}{0.97\textwidth}\small LLaVA uses 75 V* diagnostic questions; Qwen and InternVL each use 93. Every question has 24 candidate orders. Each readout projects the final prompt token through the final normalization and output head. Content L1 is the mean L1 distance of candidate distributions from their average over orders; lower is better. Variation first appears at the attention output in all three models.\end{minipage}
\end{figure*}

\subsection{Mechanistic Analysis of Permutation Sensitivity}\label{sec:diagnostic}

We conduct layer-wise analysis and controlled interventions using LLaVA-1.5-7B, Qwen2.5-VL-3B-Instruct, and InternVL3.5-4B-HF on their diagnostic subsets defined in Section~\ref{sec:subset}. Each question is evaluated under all 24 candidate permutations using the option-label-free protocol. We first trace the emergence and evolution of permutation-dependent differences across Transformer layers, and then examine the contributions of positional encoding and effective causal attention visibility.

\paragraph{Layer-wise analysis} We examine seven computational checkpoints within each Transformer block: block input, input normalization, attention output, attention residual, post-attention normalization, MLP output, and MLP residual. LLaVA has 32 blocks, while Qwen and InternVL each have 36. Including the final normalization output gives 225 checkpoints for LLaVA and 253 for each of the other two models. At each checkpoint, the hidden representation at the final prompt token is projected through the model's final RMSNorm and language modeling head to obtain candidate logits, except at the final normalization checkpoint, where only the language modeling head is applied. The resulting candidate probabilities are evaluated using Content L1. These intermediate readouts quantify candidate-distribution variation within the network.

Across all three models, the earliest measured permutation-dependent signal appears at the attention output, identifying attention as the common entry point for the structural interventions. The first-block input and input normalization have zero measured variation. Attention-output Content L1 is $\text{2.33e-3}$ for LLaVA, $\text{3.74e-3}$ for Qwen, and $\text{6.83e-3}$ for InternVL (Figure~\ref{fig:layers}(b)). Block-end trajectories show subsequent amplification and attenuation, reaching final-block Content L1 of 0.16, 0.12, and 0.15, respectively (Figure~\ref{fig:layers}(a)). The shared onset at attention extends the mechanistic observation across architectures.

\paragraph{Position-based counterfactual intervention} In causal Transformers, rearranging candidate answers changes both their positional representations and their access to preceding tokens. To examine the contribution of positional encoding, we first conduct a local intervention on rotary position embeddings (RoPE) \citep{rope}. Specifically, we reconstruct the attention computation at the final prompt query in block 0 and replace only the RoPE positions assigned to candidate keys with content-fixed positions, while keeping the query, value vectors, and attention visibility unchanged. The reconstructed attention output is added to the original block input residual and projected through the final RMSNorm and language modeling head without propagation through subsequent layers. For LLaVA, Content L1 decreases from $\text{2.95e-3}$ to $\text{9.50e-8}$, indicating that positional assignments contribute substantially to the observed local sensitivity. The same intervention reduces Content L1 from $\text{4.01e-3}$ to near zero for Qwen and from $\text{7.66e-3}$ to $\text{6.73e-6}$ for InternVL.

\paragraph{Joint intervention on position and attention visibility} We further investigate the interaction between positional encoding and causal attention visibility through a $2\times2$ controlled experiment. The first factor determines whether candidate query/key positions follow their native sequence order or a content-fixed canonical assignment. The second factor determines whether attention uses the native causal mask or a content-canonical effective visibility mask. In the latter configuration, candidate tokens are assigned content-fixed effective positions, while other prompt tokens retain their native effective positions. Attention is permitted when the key's effective position does not exceed the query's effective position. This intervention is applied throughout the language Transformer and retains cross-unit attention under a content-canonical causal order. Unlike SP-B introduced below, it does not isolate candidate units.

\begin{table*}[t]
\centering\normalsize\renewcommand{\arraystretch}{1.18}
\caption{Effects of positional assignment and attention visibility.}\label{tab:cross}
\setlength{\tabcolsep}{6pt}
\begin{tabular}{llrrrrr}\toprule
Q/K & Mask & Content L1 $\downarrow$ & OA $\uparrow$ & AA $\uparrow$ & AC $\uparrow$ & IP $\uparrow$ \\\midrule
Native & Native & 0.16 & \textbf{42.67} & \textbf{38.94} & 17.33 & 49.33\\
Canonical & Native & 0.29 & 32.00 & 28.72 & 6.67 & 20.00\\
Native & Fixed & 0.24 & 29.33 & 30.22 & 9.33 & 26.67\\
Canonical & Fixed & $\mathbf{1.19e-2}$ & 32.00 & 32.39 & \textbf{30.67} & \textbf{96.00}\\
\bottomrule\end{tabular}

\par\smallskip\begin{minipage}{0.97\textwidth}\small
LLaVA is evaluated on 75 V* diagnostic questions, with 24 orders per question. Canonical Q/K assigns positions by candidate content rather than input order. Fixed mask sets visibility using this content-based order. Native uses the original positions or causal mask. Content L1 measures variation in content-aligned candidate distributions; lower is better. It is unscaled; the other metrics are percentages. OA: original-order accuracy; AA: mean accuracy across permutations; AC: accuracy requiring every order to be correct; IP: the share of questions with the same prediction across all orders. Bold marks the best value in each column.
\end{minipage}
\end{table*}

For LLaVA, jointly controlling positional assignments and effective visibility raises IP from 49.33\% to 96.00\% and reduces Content L1 from 0.16 to $\text{1.19e-2}$ (Table~\ref{tab:cross}). Either isolated intervention increases Content L1 relative to native inference. This contrast identifies position and visibility as coupled design variables, motivating the coordinated shared-position and attention-mask structure in Section~\ref{sec:shared}. The corresponding Qwen and InternVL local-intervention and position--visibility controls are reported in the submitted Supplementary Material, Section S5 (Figure S3 and Table S1). Joint control achieves 94.62\% IP in both models.

\subsection{Shared Positions with Blocked Cross-Unit Attention}\label{sec:shared}

The analysis in Section~\ref{sec:diagnostic} suggests that eliminating order dependence in both positional encoding and attention visibility should remove permutation bias in exact arithmetic. Based on this reasoning and existing set-based positional designs \citep{positional_overload}, we introduce shared positions with blocked cross-unit attention (SP-B). SP-B assigns shared positional coordinates to exchangeable units and prevents attention between different units. These changes are intended to make predictions invariant to the order of those units. Section~\ref{sec:numerics} tests this expectation and examines the effects of finite precision.

\paragraph{Shared positional encoding} Following Section~\ref{sec:formulation}, let $E_i$ contain $\ell_i$ tokens after multimodal expansion. For scalar positions, all units share a start $\rho_0$, and token $t$ within unit $E_i$ has position

\begin{equation}\rho(E_i,t)=\rho_0+t,\qquad 0\leq t<\ell_i.\label{eq:positions}\end{equation}

The prefix retains its original positions, while the suffix begins at $\rho_0+\max_i\ell_i$. This preserves internal token order while removing offsets caused solely by unit reordering. Units of different lengths require neither padding nor truncation because only their starting positions are shared.

\begin{figure*}[tp]\centering\includegraphics[width=\textwidth,height=0.60\textheight,keepaspectratio]{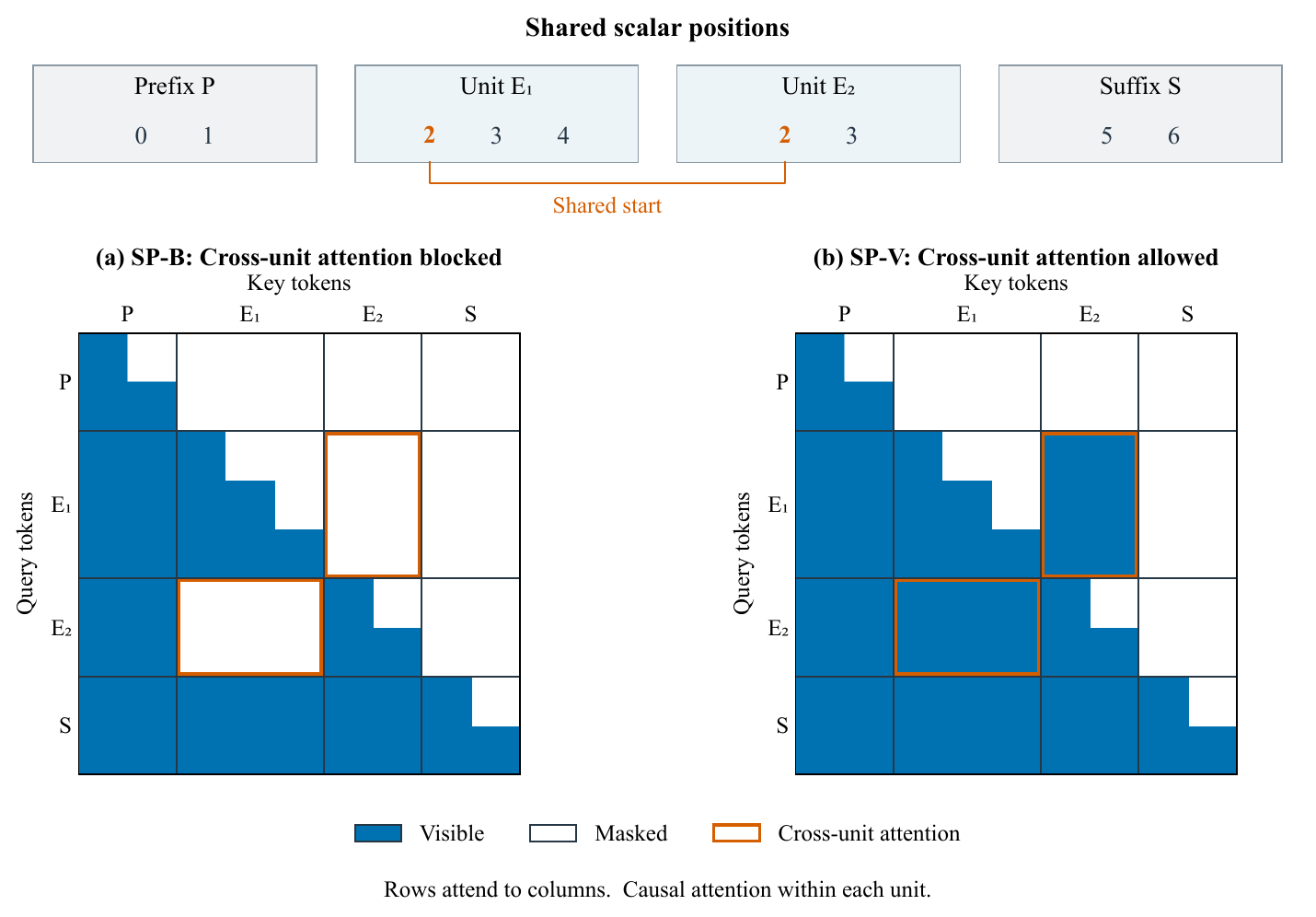}\caption{Shared-position attention configurations: (a) blocked cross-unit attention (SP-B); (b) bidirectional cross-unit attention (SP-V).}\label{fig:method}
\par\smallskip\begin{minipage}{0.97\textwidth}\small Exchangeable units share a starting position, even when their lengths differ. SP-B blocks attention between units. SP-V allows attention in both directions. Both retain causal order within each unit.\end{minipage}
\end{figure*}

\paragraph{Blocked cross-unit attention} Shared positional encoding alone does not eliminate order-dependent causal visibility. We therefore modify the attention mask to prevent information exchange between different units while preserving causal attention within each unit. For a query token $u\in E_i$ and a key token $v\in E_j$, the visibility rule is defined as

\begin{equation}B_{\mathrm B}(u,v)=\begin{cases}\mathbb I[v\preceq u],&i=j,\\0,&i\ne j.\end{cases}\label{eq:blocked}\end{equation}

where $v\preceq u$ denotes the local causal order (including self-attention) within the same unit. Each unit can attend to the fixed prefix and its own preceding tokens, but not to other exchangeable units or the suffix. The prefix retains causal attention, whereas the suffix can attend to the prefix, all exchangeable units, and its own preceding tokens. As illustrated in Figure~\ref{fig:method}(a), this configuration preserves causal diagonal blocks while masking all off-diagonal cross-unit attention blocks.

The binary visibility $B$ is converted to an additive attention mask:
\begin{equation}M_{uv}=\begin{cases}0,&B(u,v)=1,\\-\infty,&B(u,v)=0.\end{cases}\label{eq:additive-mask}\end{equation}
The prefix and suffix relationships described above complete this mask outside the exchangeable-unit blocks.

Mapping to multimodal tokens. To apply SP-B across the four permutation scenarios, exchangeable units are defined before inference and mapped to their complete token spans after multimodal expansion. In text-option reordering, each unit contains an entire candidate-answer line, including its list marker and newline. In image reordering, each unit includes the image placeholder, expanded visual tokens, associated delimiters, and trailing newline. For modality exchange, the two units correspond to the complete image block and the textual question block, including candidate answers and instructions. In image--text demonstration reordering, each unit contains a complete image--text example with its associated formatting. The assistant marker remains in the fixed suffix. Consequently, positional sharing and attention masking are applied to complete semantic units rather than isolated answer tokens or visual patches.

For LLaVA-1.5 and InternVL3.5, shared positions are assigned to expanded token spans using scalar language-side indices. Qwen2.5-VL instead uses three-axis coordinates. Let $k$ denote a token index in the expanded sequence, $r_a(k)$ its native coordinate on axis $a$, and $s_i$ the start index of unit $E_i$ in the displayed sequence. Define $s_{\mathrm{first}}=\min_i s_i$ and $b_a=r_a(s_{\mathrm{first}})$. The shared-position mapping is

\begin{equation}\rho_a(k)=b_a+r_a(k)-r_a(s_i),\qquad s_i\leq k<s_i+\ell_i.\label{eq:mrope}\end{equation}

This mapping preserves the relative positional coordinates within each unit, including the spatial structure of visual tokens, while assigning a common anchor to all exchangeable units within the same forward pass. The prefix retains its original coordinates, and the suffix is positioned after the maximum assigned coordinate across all units and axes, preserving its native axis-wise offsets. The anchor is computed separately from the native position tensor for each input order. SP-B modifies language-side positions and attention visibility while retaining native image preprocessing and vision encoding.

SP-B provides a permutation-compatible structure that aggregates independently encoded units through the fixed suffix. Tasks requiring direct interaction between units can use the bidirectional configuration in Section~\ref{sec:bidir}. Section~\ref{sec:numerics} will test the effectiveness of SP-B on the diagnostic subset.

\subsection{The Role of Shared Positions and Numerical Precision}\label{sec:numerics}

Based on the analysis in Section~\ref{sec:diagnostic}, the SP-B design in Section~\ref{sec:shared} should eliminate permutation bias in exact arithmetic. In practice, however, input permutations can still change the order of floating-point operations. With finite precision, different computation orders may produce numerical differences and affect predictions. We therefore test whether SP-B achieves permutation invariance and examine the roles of numerical precision and computation order.

\paragraph{Experimental configurations} We compare the native Transformer structure and SP-B under three configurations: ordinary BF16 inference, ordinary FP32 inference, and BF16 inference with canonical attention computation order. The canonical configuration evaluates complete candidate sequences in a fixed order determined by their content hashes. We reorder the corresponding query rows, key/value columns, positional inputs, and attention masks together. After attention, we restore the outputs to their original tensor order. This fixes the attention computation order without changing the input content, logical positions, or attention visibility.

\begin{table*}[t]
\centering\normalsize\renewcommand{\arraystretch}{1.18}
\caption{Effects of numerical precision and attention computation order.}\label{tab:numerical}
\setlength{\tabcolsep}{6pt}
\begin{tabular}{lllrrrrr}\toprule
Structure & Precision & Attention order & Content L1 $\downarrow$ & OA $\uparrow$ & AA $\uparrow$ & AC $\uparrow$ & IP $\uparrow$ \\\midrule
Native & BF16 & Ordinary & 0.16 & \textbf{42.67} & 38.94 & 17.33 & 49.33\\
SP-B & BF16 & Ordinary & $\text{1.66e-2}$ & 38.67 & 39.56 & \textbf{38.67} & 93.33\\
Native & FP32 & Ordinary & 0.16 & \textbf{42.67} & \textbf{39.61} & 18.67 & 48.00\\
SP-B & FP32 & Ordinary & $\text{7.65e-7}$ & 38.67 & 38.67 & \textbf{38.67} & \textbf{100}\\
Native & BF16 & Canonical & 0.16 & \textbf{42.67} & 38.83 & 16.00 & 46.67\\
SP-B & BF16 & Canonical & \textbf{0.00} & 38.67 & 38.67 & \textbf{38.67} & \textbf{100}\\
\bottomrule\end{tabular}

\par\smallskip\begin{minipage}{0.97\textwidth}\small
LLaVA is evaluated on the same 75 V* diagnostic questions and 24 orders per question. SP-B shares positions and blocks attention between candidate units. Canonical attention fixes the computation order by content, while keeping logical positions and visibility unchanged. Content L1 measures variation in candidate distributions and is unscaled; lower is better. The other metrics are percentages. OA: original-order accuracy; AA: mean accuracy across permutations; AC: accuracy requiring every order to be correct; IP: the share of questions with the same prediction across all orders. Bold marks the best value in each column.
\end{minipage}
\end{table*}

\paragraph{Can SP-B remove permutation bias?} Under ordinary BF16 inference, SP-B raises LLaVA's IP from 49.33\% to 93.33\% and reduces Content L1 from 0.16 to $\text{1.66e-2}$ (Table~\ref{tab:numerical}). Thus, SP-B greatly reduces permutation sensitivity, but some predictions still change across orders. With FP32, IP reaches 100\% and Content L1 falls to $\text{7.65e-7}$. All evaluated questions then receive consistent predictions across permutations, although tiny differences remain in their candidate probabilities. Higher precision therefore enables prediction invariance under SP-B on this diagnostic subset.

\paragraph{Why do numerical differences remain?} Native inference remains sensitive to permutations under both BF16 and FP32. Higher precision alone therefore does not resolve the structural problem. We next keep SP-B and BF16 fixed and change only the attention computation order. Canonical attention raises IP to 100\% and reduces measured Content L1 to zero. SP-B aligns logical positions and attention visibility across permutations, but ordinary attention can still accumulate values in different orders. Floating-point addition is not associative, so these orders can produce different results. Fixing the attention computation order removes the residual variation in this experiment. The same order control leaves Native Content L1 at 0.16, confirming that computation-order control does not replace structural adaptation.

In addition, Qwen2.5-VL and InternVL show similar results in the Supplementary Material.

These experiments distinguish two sources of permutation sensitivity: structural dependence on positions and visibility, and residual numerical differences caused by computation order. For the subsequent deployment evaluations, we use ordinary BF16 inference. We do not enforce canonical attention order or require separate encoding of each image. This setting is more general and applicable to common test scenarios.

\subsection{Extension to Bidirectional Cross-Unit Attention}\label{sec:bidir}

Some exchangeable units may require direct information exchange, as in image--question interaction. Shared positions with bidirectional cross-unit attention (SP-V) retain the positional assignments of Section~\ref{sec:shared} while enabling symmetric cross-unit attention.

\paragraph{Bidirectional cross-unit attention} SP-V modifies only the off-diagonal attention blocks between exchangeable units, replacing the blocked connections in SP-B with bidirectional visibility. For a query token $u\in E_i$ and a key token $v\in E_j$, the visibility rule becomes

\begin{equation}B_{\mathrm V}(u,v)=\begin{cases}\mathbb I[v\preceq u],&i=j,\\1,&i\ne j.\end{cases}\label{eq:visible}\end{equation}

where $v\preceq u$ denotes the local causal order within the same unit. The prefix and suffix retain the same attention relationships as in SP-B: each unit can attend to the prefix, the prefix cannot attend to subsequent units, and the suffix can attend to all preceding tokens. As illustrated in Figure~\ref{fig:method}(b), SP-V preserves causal attention within individual units while enabling bidirectional interactions between different units. Each Transformer layer computes these interactions from representations produced by the preceding layer.

\paragraph{Conditional permutation invariance} Both SP-B and SP-V are designed to preserve permutation symmetry at the structural level. Let $\Pi$ denote a permutation matrix that reorders complete expanded units while leaving the prefix and suffix unchanged. Suppose that (1) the input token representations are permuted consistently without additional content changes, (2) positional coordinates are permuted accordingly, (3) the attention mask satisfies $M_\pi=\Pi M\Pi^\top$, and (4) computations are performed using exact arithmetic with deterministic token-shared operations. Under these assumptions, the self-attention operator $\mathcal A$ is permutation-equivariant:

\begin{equation}
\mathcal A(\Pi Q,\Pi K,\Pi V;\Pi M\Pi^\top)=\Pi\mathcal A(Q,K,V;M).
\label{eq:equivariance}\end{equation}
Here $Q$ and $K$ include their positional transformation, and $M$ is the additive mask. Token-shared projections, normalization, MLPs, and residual connections preserve permutation equivariance, so the property extends across Transformer layers. A fixed suffix representation consequently yields a permutation-invariant prediction under deterministic, content-aligned tie-breaking. Both blocked and bidirectional cross-unit attention satisfy the required mask symmetry, provided that positional assignments and expanded token representations correspond consistently across permutations.

\section{Experimental Evaluation and Accuracy Recovery}\label{sec:experiments}

In this section, we evaluate the proposed methods across the four permutation scenarios. We first examine the effectiveness of the SP structures (SP-B and SP-V) under native inference settings, and then evaluate the proposed attention-only distillation framework for capability recovery. Finally, we compare our complete method with majority voting and fine-tuning to show its overall performance and robustness.

\subsection{Experimental Setup}\label{sec:setup}

\paragraph{Models} We evaluate three representative MLLMs: LLaVA-1.5-7B \citep{llava15}, Qwen2.5-VL-3B-Instruct \citep{qwen25vl}, and InternVL3.5-4B-HF \citep{internvl35}.

\paragraph{Datasets and permutation scenarios} We construct eight evaluation settings across the four permutation scenarios defined in Section~\ref{sec:units}, as summarized in Table~\ref{tab:tasks}. For text-option reordering and modality exchange, we use V*Bench (V*) \citep{vstar_data} and the A-OKVQA validation set \citep{aokvqa_data}. Image reordering is evaluated using a subset of 153 symmetric two-book comparison questions from the standard OCR-VQA task in MLBench \citep{ocrvqa_data,future_attention} and 1,000 two-image yes/no questions reconstructed from a subset of POPE \citep{pope}, referred to as POPE below. Image--text demonstration reordering is evaluated using 96 target images from CelebA-HQ \citep{celebahq_data} with four fixed demonstrations, and 1,161 target questions from MMBench \citep{mmbench_data} with three fixed demonstrations.

\begin{table*}[t]
\centering\normalsize\renewcommand{\arraystretch}{1.18}\caption{Evaluation datasets and permutation configurations.}\label{tab:tasks}
\begin{tabular}{llrrr}\toprule
Scene & Dataset & Questions & Orders/question & Records\\\midrule
Text & V* &191&2 or 24&2,736\\
Text & A-OKVQA val &1,145&24&27,480\\
Image & OCR-VQA &153&2&306\\
Image & POPE &1,000&2&2,000\\
Image--text & CelebA-HQ &96&24&2,304\\
Image--text & MMBench &1,161&6&6,966\\
Modality & V* &191&2&382\\
Modality & A-OKVQA val &1,145&2&2,290\\
\midrule
Total &&5,082&&44,464\\\bottomrule
\end{tabular}
\par\smallskip\begin{minipage}{0.97\textwidth}\small
Records count all evaluated orders per model. The total counts question--setting pairs, since text and modality exchange reuse the same questions. OCR-VQA is the 153-question two-image subset in MLBench. POPE uses constructed two-image questions. CelebA-HQ and MMBench use four and three fixed demonstrations, respectively; these are excluded from the target-question counts.
\end{minipage}
\end{table*}

Predictions are evaluated using OA, AA, AC, and IP (Section~\ref{sec:metrics}). Task-specific scoring protocols are provided in the Supplementary Material, Section S1.3.

\paragraph{Experimental comparisons} We evaluate frozen shared-position inference to assess model performance under the SP structure before adaptation. We compare performance before and after knowledge distillation to assess whether training recovers predictive accuracy under the SP structure. Across all three models, we compare our framework with majority voting and supervised fine-tuning (FT) in terms of predictive accuracy, permutation robustness, and inference requirements.

\subsection{Permutation Robustness Across Four Scenarios}\label{sec:frozen}
\begin{figure*}[tp]\centering\includegraphics[width=\textwidth,height=0.60\textheight,keepaspectratio]{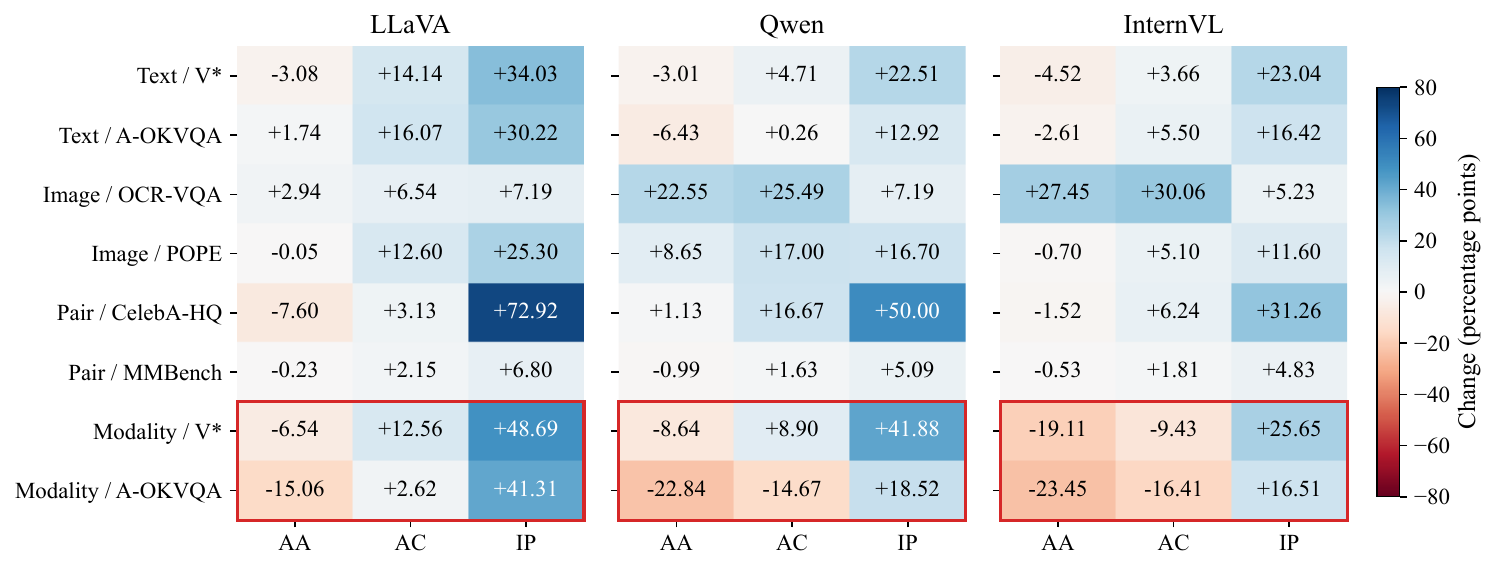}\caption{Changes from native inference to frozen SP-B. Columns show AA, AC, and IP; rows show the eight settings. SP-B blocks cross-unit attention in every scenario. Red outlines mark modality exchange.}\label{fig:frozen}
\par\smallskip\begin{minipage}{0.97\textwidth}\small Values are frozen SP-B minus native inference, in percentage points; positive values mean an increase. AA is mean accuracy across orders. AC requires every order to be correct. IP measures prediction consistency across orders. No weights are updated.\end{minipage}
\end{figure*}

Frozen SP-B improves Invariant Predictions (IP) in all 24 model--setting groups compared with native inference (Figure~\ref{fig:frozen}). It also improves All-permutation Correct Accuracy (AC) in all 18 groups for text-option, image, and image--text demonstration reordering. For modality exchange, AC improves in some groups but decreases in others. Average Accuracy (AA) increases in 6 of the 24 groups and decreases in the remaining 18.

The red boxes mark the modality-exchange rows. This is the only scenario in which SP-B does not improve AC in every group. One possible reason is that SP-B blocks attention between the image and question, preventing direct information exchange between them. We therefore compare SP-V with SP-B for modality exchange in Section~\ref{sec:visibility}.

\subsection{Effects of Bidirectional Cross-Unit Attention in Modality Exchange}\label{sec:visibility}
\begin{figure*}[tp]\centering\includegraphics[width=\textwidth,height=0.60\textheight,keepaspectratio]{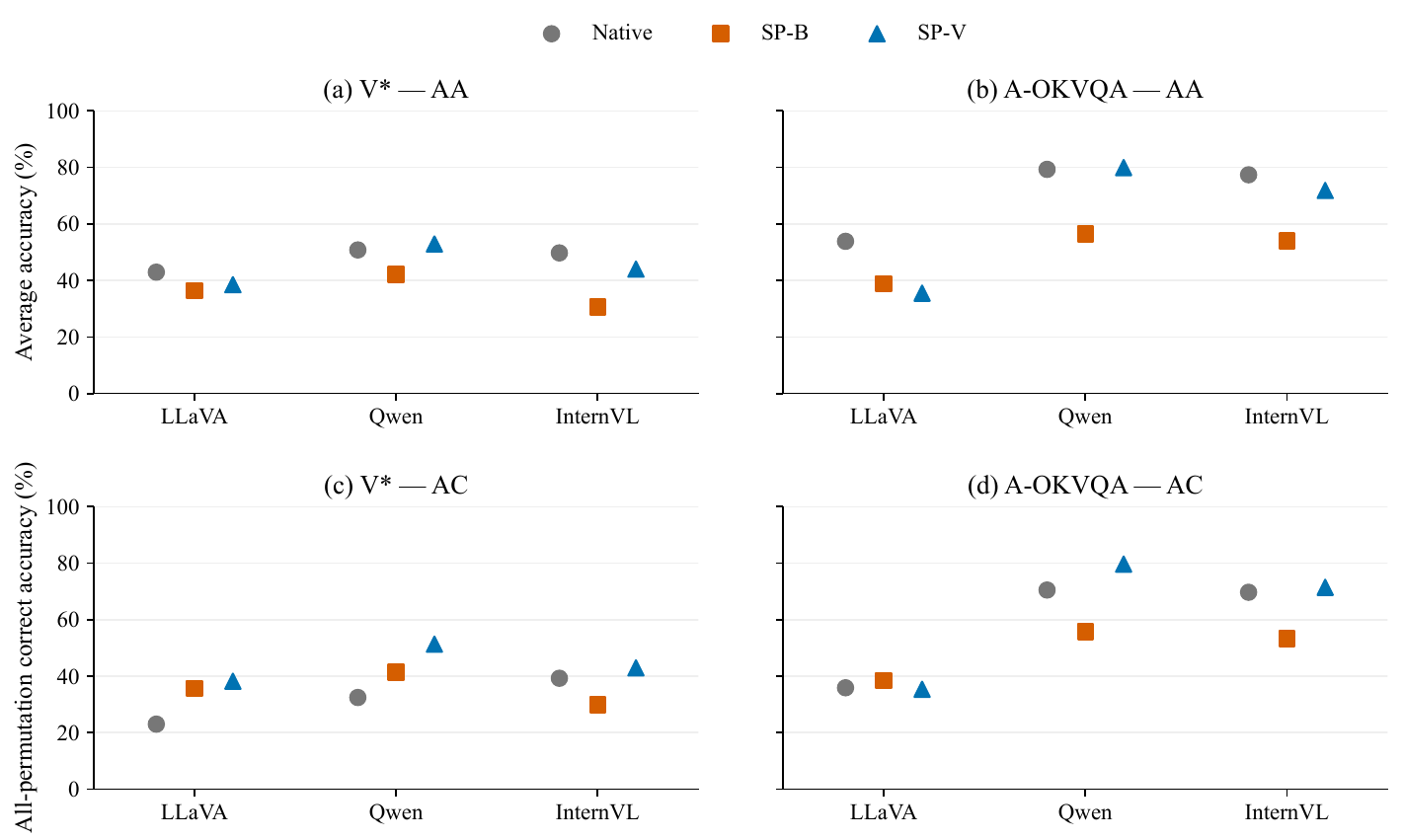}\caption{Frozen-model modality exchange: AA (top) and AC (bottom) under native inference, SP-B, and SP-V.}\label{fig:visibility}
\par\smallskip\begin{minipage}{0.97\textwidth}\small No weights are updated. Native uses the original model. SP-B blocks attention between the image and question blocks; SP-V allows attention in both directions. AA is mean accuracy across orders. AC requires every order to be correct. SP-V increases both metrics over SP-B in five of six groups.\end{minipage}
\end{figure*}

Figure~\ref{fig:visibility} compares AA and AC under frozen native inference, SP-B, and SP-V for modality exchange. SP-V allows bidirectional attention between the image and question. Compared with SP-B, it improves both metrics in five of the six model--dataset groups. On A-OKVQA, AA increases from 56.46\% to 79.91\% for Qwen and from 53.93\% to 71.83\% for InternVL, while for LLaVA the two SP structures yield close AA. Overall, these results support allowing direct attention between the image and question under shared-position inference.

We therefore use SP-V for modality exchange and SP-B for the other three scenarios in the subsequent adaptation and evaluation experiments.

\subsection{Knowledge Distillation for Structural Adaptation}\label{sec:training}

\begin{figure*}[tp]\centering\includegraphics[width=\textwidth,height=0.82\textheight,keepaspectratio]{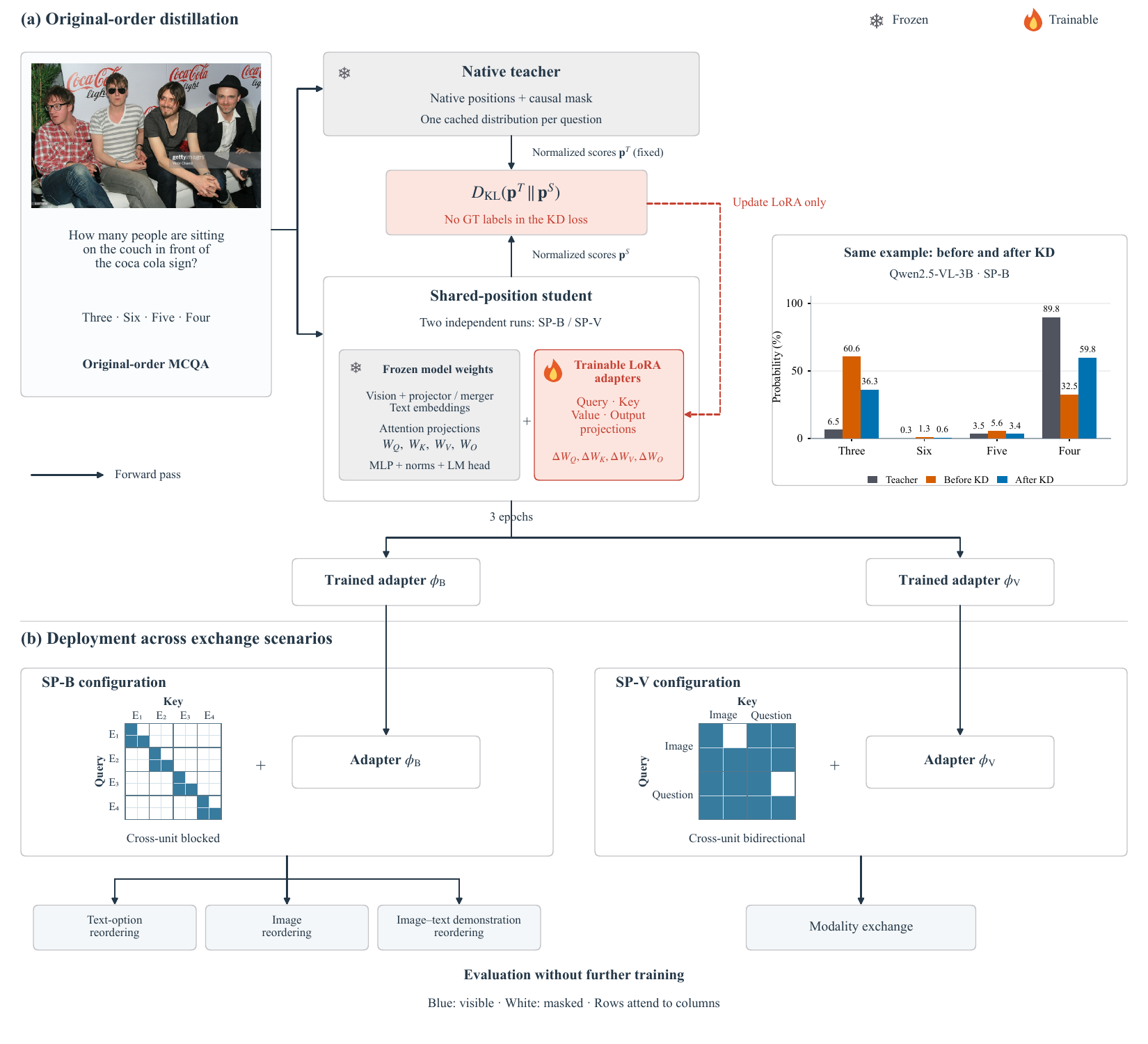}\caption{Distillation and deployment of shared-position adapters.}\label{fig:distill}
\par\smallskip\begin{minipage}{0.97\textwidth}\small KD means knowledge distillation. Only attention LoRA is trained; the KD loss uses teacher distributions without ground-truth labels. In the SEED-Bench example, Qwen changes its preferred count from ``Three to ``Four after three epochs, matching the teacher. The two adapters are trained independently and used with their own attention masks in (b).\end{minipage}
\end{figure*}

Distillation aligns the native model's candidate distribution with the shared-position inference structure. For each training question, the teacher (the frozen native model) runs only on the original order and caches one candidate distribution as the target. The student then adapts only the attention projections to match this distribution, while the pretrained backbone is kept fixed. Figure~\ref{fig:distill} summarizes the training and deployment design.

\paragraph{Teacher--student framework} For each MLLM, the frozen native model serves as the teacher, using its original positional encoding, causal attention mask, and original candidate ordering. The student is initialized from the same pretrained checkpoint but employs shared positional encoding and modified attention visibility. Both configurations use original-order, option-label-free MCQA for distillation. SP-B assigns shared positions to candidate-answer blocks and blocks attention between them; SP-V assigns shared positions to the image and complete question blocks and permits bidirectional attention between them. Both teacher and student receive the same image, question, and candidate contents. For each training question, we cache a single candidate-score distribution from the teacher under the original input order, without evaluating additional permutations. The student learns the native model's original-order candidate preferences under the shared-position structure.

\paragraph{Knowledge-distillation objective} For candidate $c_j=(c_{j1},\ldots,c_{jL_j})$ containing $L_j$ tokens, we compute its length-normalized conditional log-probability under the teacher or student as

\begin{equation}s_j^{(u)}=\frac{1}{L_j}\sum_{t=1}^{L_j}\log P_u(c_{jt}\mid X,c_{j,<t}),\quad u\in\{T,S\}.\label{eq:scores}\end{equation}

where $T$ and $S$ denote the teacher and student, respectively. The candidate scores are normalized into probability distributions using temperature $\tau$:

\begin{equation}p_j^{T}=\frac{\exp(s_j^{(T)}/\tau)}{\sum_k\exp(s_k^{(T)}/\tau)},\qquad p_j^{S}=\frac{\exp(s_j^{(S)}/\tau)}{\sum_k\exp(s_k^{(S)}/\tau)}.\label{eq:distributions}\end{equation}

Let $\mathbf p^T=(p_j^T)_j$ and $\mathbf p^S=(p_j^S)_j$ denote these candidate distributions. The student minimizes their Kullback--Leibler divergence:

\begin{equation}\ell_{\mathrm{KD}}(X)=\tau^2\sum_j p_j^{T}\log\frac{p_j^{T}}{p_j^{S}},\quad\tau=1.\label{eq:kl}\end{equation}

The training objective is the mean $\mathcal L_{\mathrm{KD}}=N_{\mathrm{tr}}^{-1}\sum_{n=1}^{N_{\mathrm{tr}}}\ell_{\mathrm{KD}}(X_n)$ over training questions. The teacher and student distributions normalize length-averaged log scores over the supplied candidates. Optimization uses this KL objective alone. Ground-truth annotations support question selection, validation scoring, and checkpoint selection.

\paragraph{Parameter-efficient adaptation} To adapt the student while preserving most pretrained parameters, we introduce low-rank adaptation (LoRA) \citep{lora} into the query, key, value, and output projections of the language-model attention modules. Each adapted weight matrix is expressed as

\begin{equation}W^{\prime}=W+\frac{\alpha}{r}B_{\mathrm L}A_{\mathrm L},\qquad r=16,\quad\alpha=32.\label{eq:lora}\end{equation}

For $W\in\mathbb R^{d_{\mathrm{out}}\times d_{\mathrm{in}}}$, $A_{\mathrm L}\in\mathbb R^{r\times d_{\mathrm{in}}}$ and $B_{\mathrm L}\in\mathbb R^{d_{\mathrm{out}}\times r}$ are trainable matrices. The effective weight update is $\Delta W=(\alpha/r)B_{\mathrm L}A_{\mathrm L}$. In Figure~\ref{fig:distill}, $\phi_{\mathrm B}$ and $\phi_{\mathrm V}$ denote the LoRA parameters learned in the independent SP-B and SP-V runs, respectively. LoRA dropout is set to zero. All original model weights, MLP modules, visual encoders, multimodal projectors, and embedding layers remain frozen. Thus, adaptation is restricted to attention projections while the shared-position inference structure remains unchanged. Since LoRA updates are token-shared linear transformations, they preserve the conditional permutation-equivariance properties discussed in Section~\ref{sec:bidir} under the corresponding idealized assumptions.

\paragraph{Training protocol} Independent SP-B and SP-V adapters start from each native checkpoint. Fixed SEED-Bench \citep{seedbench} lists contain 1,591 SP-B training questions, 1,592 SP-V training questions, and 1,000 validation questions per configuration. Training uses original orders only. Validation OA selects epoch three for all six adapters. Question selection incorporates LLaVA native/shared-position behavior, and the resulting lists are shared across all three models.

\begin{table*}[t]
\centering\small\renewcommand{\arraystretch}{1.16}
\caption{Training configurations.}\label{tab:training_config}
\begin{tabular}{lll}\toprule
Configuration & Knowledge distillation (KD) & Supervised fine-tuning (FT)\\\midrule
Models & LLaVA, Qwen, InternVL & LLaVA, Qwen, InternVL\\
Training questions (SP-B / SP-V) & 1,591 / 1,592 & Same lists\\
Validation questions & 1,000 & Same list\\
Orders per question per epoch & One fixed original order & One randomly sampled order\\
Objective & Teacher candidate-distribution KL & Full-sequence text cross-entropy\\
Updated parameters & Attention Q/K/V/O LoRA & All language-model parameters\\
Visual modules & Frozen & Frozen\\
Optimizer & AdamW & Adafactor\\
Learning rate & $2\times10^{-5}$ & $5\times10^{-6}$\\
Microbatch / gradient accumulation & 1 / 16 & 1 / 64\\
Reported checkpoint & Epoch 3 for all six adapters & Best validation OA within 3 epochs\\
\bottomrule\end{tabular}
\par\smallskip\begin{minipage}{0.97\textwidth}\small
Both methods use the same SEED-Bench training and validation question lists. KD means knowledge distillation; FT means supervised fine-tuning. Both use the corresponding SP-B or SP-V structure. SP-B and SP-V runs start independently from native weights. FT selects the highest validation OA within three epochs. SP-B selects epoch 3 for all models; SP-V selects epochs 3, 2, and 1 for LLaVA, Qwen, and InternVL, respectively. Detailed LoRA, warmup, and random-seed settings are provided in the Supplementary Material.
Their losses, trainable parameters, and order sampling differ.
\end{minipage}\end{table*}

\paragraph{Deployment and transfer} As illustrated in Figure~\ref{fig:distill}, the adapter trained with candidate-block SP-B is subsequently evaluated on text-option, image, and image--text demonstration reordering without additional scene-specific training. The independently trained SP-V adapter is used for modality exchange. During deployment, only the adapted student is required. Neither teacher inference nor permutation-based voting is necessary. The resulting capability recovery and cross-scenario transfer performance are evaluated in Section~\ref{sec:trained}.

\subsection{Accuracy Recovery and Cross-Scenario Transfer}\label{sec:trained}
SP + KD improves All-permutation Correct Accuracy over native inference in all 24 model--setting comparisons, with gains of 1.46--32.03 percentage points (Table~\ref{tab:recovery}). Across the 24 groups, the macro-average gap between AA and AC narrows from 9.98 to 0.59 percentage points, showing that correct predictions become substantially more stable across input permutations. Macro-average IP increases from 73.20\% for native inference to 97.44\% for frozen SP and 98.20\% for SP + KD. SP denotes SP-V for modality exchange and SP-B elsewhere.

\begin{figure*}[tp]\centering\includegraphics[width=0.98\textwidth,height=0.76\textheight,keepaspectratio]{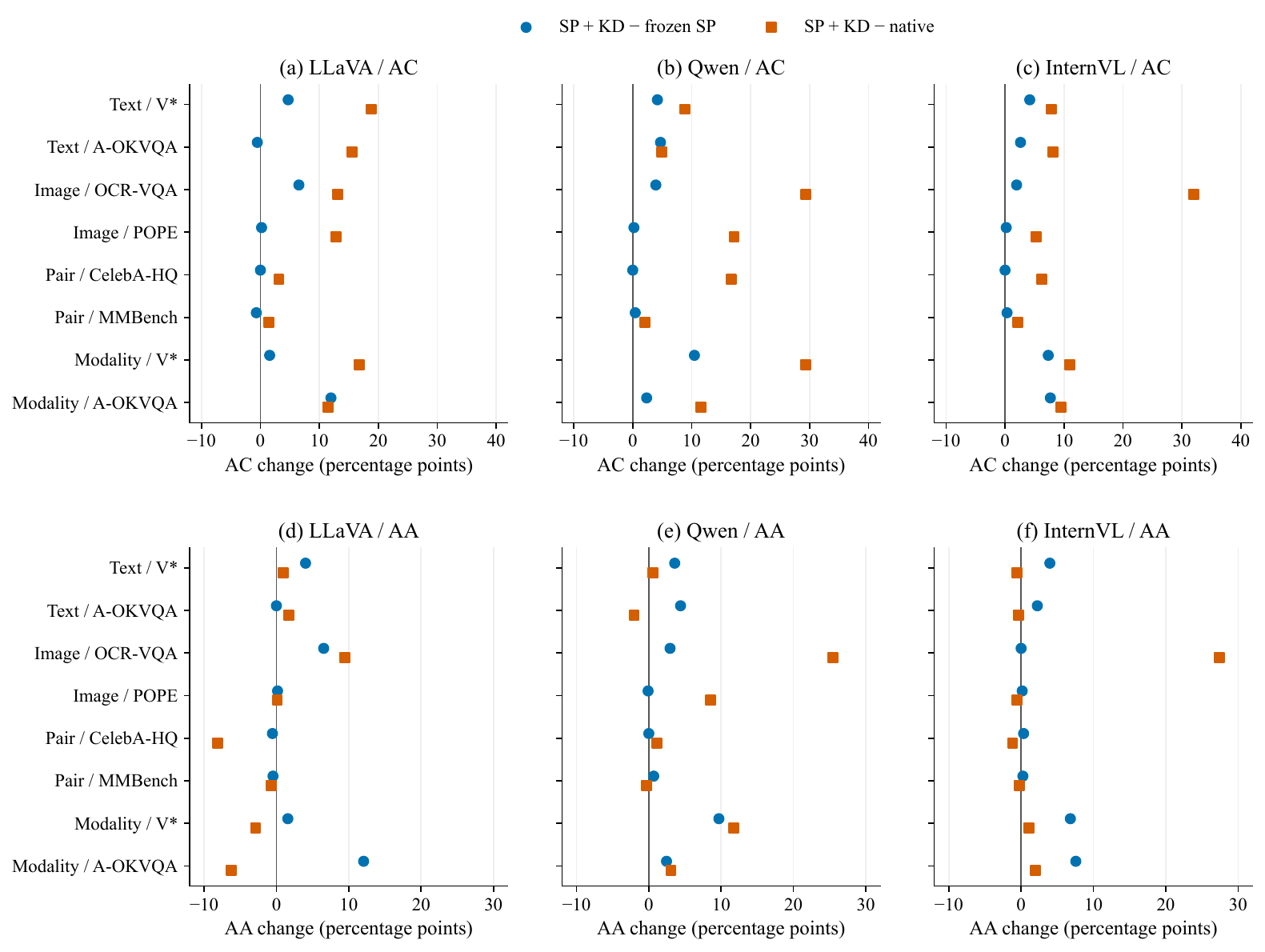}\caption{Accuracy changes after adaptation: AC (top row) and AA (bottom row) for LLaVA, Qwen, and InternVL (left to right). Blue circles show SP + KD minus frozen SP; orange squares show SP + KD minus native inference. Differences are in percentage points. SP is SP-V for modality exchange and SP-B elsewhere.}\label{fig:recovery}
\par\smallskip\begin{minipage}{0.97\textwidth}\small KD is knowledge distillation; all models use epoch-3 weights. AA is mean accuracy across orders. AC requires every order to be correct. The blue contrast measures the change due to KD. The orange contrast measures the full method against native inference. The full method increases AC in all 24 groups.\end{minipage}
\end{figure*}

\paragraph{Recovery under the modified structure} Modality adaptation increases A-OKVQA AA by 12.05 percentage points for LLaVA and 7.56 points for InternVL. On Qwen V*, it raises modality AA from 52.88\% to 62.57\%. These comparisons isolate the contribution of KD relative to frozen shared-position inference. Figure~\ref{fig:recovery} separates this adaptation gain from the complete framework's improvement over native inference.

\paragraph{Transfer to image and demonstration reordering} Adapters trained with candidate-block SP-B transfer to image and demonstration reordering without additional scenario-specific training. The resulting models improve AC over native inference in all 12 transfer settings. 

\paragraph{Accuracy and consistency} AC measures correctness sustained across every valid input ordering, whereas AA measures correctness averaged over those orderings. Their close alignment after adaptation means that average performance increasingly reflects predictions that remain correct under permutation. 
\begin{table*}[t]
\centering\small\renewcommand{\arraystretch}{1.13}
\caption{Accuracy and invariance before and after adaptation.}\label{tab:recovery}
\setlength{\tabcolsep}{4pt}
\begin{tabular}{ll|c|c|c}\toprule
Model & Setting & Native & Frozen SP & SP + KD\\\midrule
LLaVA & Text: V* & 46.62 / 28.27 / 61.26 & 43.54 / 42.41 / 95.29 & \textbf{47.56} / \textbf{47.12} / \textbf{98.95}\\
 & Text: A-OKVQA & 68.98 / 54.06 / 67.42 & \textbf{70.72} / \textbf{70.13} / 97.64 & 70.71 / 69.61 / \textbf{97.73}\\
 & Image: OCR-VQA & 74.84 / 71.24 / 92.16 & 77.78 / 77.78 / 99.35 & \textbf{84.31} / \textbf{84.31} / \textbf{100.00}\\
 & Image: POPE & 77.75 / 64.50 / 73.50 & 77.70 / 77.10 / 98.80 & \textbf{77.85} / \textbf{77.30} / \textbf{98.90}\\
 & Pair: CelebA-HQ & \textbf{25.87} / 14.58 / 20.83 & 18.27 / \textbf{17.71} / 93.75 & 17.71 / \textbf{17.71} / \textbf{98.96}\\
 & Pair: MMBench & \textbf{65.98} / 62.79 / 91.13 & 65.75 / \textbf{64.94} / \textbf{97.93} & 65.27 / 64.25 / 97.59\\
 & Modality: V* & \textbf{42.93} / 23.04 / 48.69 & 38.48 / 38.22 / \textbf{99.48} & 40.05 / \textbf{39.79} / 98.95\\
 & Modality: A-OKVQA & \textbf{53.84} / 35.90 / 57.64 & 35.50 / 35.37 / \textbf{99.74} & 47.55 / \textbf{47.34} / 99.39\\
 & Macro-average & \textbf{57.10} / 44.30 / 64.08 & 53.47 / 52.96 / 97.75 & 56.38 / \textbf{55.93} / \textbf{98.81}\\
\midrule
Qwen & Text: V* & 61.71 / 51.83 / 73.30 & 58.70 / 56.54 / \textbf{95.81} & \textbf{62.28} / \textbf{60.73} / \textbf{95.81}\\
 & Text: A-OKVQA & \textbf{84.60} / 76.51 / 83.06 & 78.17 / 76.77 / 95.98 & 82.56 / \textbf{81.48} / \textbf{97.73}\\
 & Image: OCR-VQA & 61.44 / 57.52 / 90.85 & 83.99 / 83.01 / 98.04 & \textbf{86.93} / \textbf{86.93} / \textbf{100.00}\\
 & Image: POPE & 66.25 / 57.00 / 81.50 & \textbf{74.90} / 74.00 / 98.20 & 74.80 / \textbf{74.20} / \textbf{98.80}\\
 & Pair: CelebA-HQ & 46.74 / 30.21 / 42.71 & \textbf{47.87} / \textbf{46.88} / 92.71 & \textbf{47.87} / \textbf{46.88} / \textbf{94.79}\\
 & Pair: MMBench & \textbf{80.20} / 76.92 / 92.76 & 79.21 / 78.55 / \textbf{97.85} & 79.89 / \textbf{78.98} / \textbf{97.85}\\
 & Modality: V* & 50.79 / 32.46 / 55.50 & 52.88 / 51.31 / 96.34 & \textbf{62.57} / \textbf{61.78} / \textbf{97.38}\\
 & Modality: A-OKVQA & 79.30 / 70.48 / 79.56 & 79.91 / 79.65 / \textbf{99.21} & \textbf{82.36} / \textbf{82.01} / 98.95\\
 & Macro-average & 66.38 / 56.62 / 74.90 & 69.45 / 68.34 / 96.77 & \textbf{72.41} / \textbf{71.62} / \textbf{97.66}\\
\midrule
InternVL & Text: V* & \textbf{52.36} / 43.46 / 73.82 & 47.84 / 47.12 / 96.86 & 51.81 / \textbf{51.31} / \textbf{97.38}\\
 & Text: A-OKVQA & \textbf{81.59} / 72.58 / 81.40 & 78.98 / 78.08 / 97.82 & 81.22 / \textbf{80.70} / \textbf{99.13}\\
 & Image: OCR-VQA & 52.94 / 48.37 / 90.85 & \textbf{80.39} / 78.43 / 96.08 & \textbf{80.39} / \textbf{80.39} / \textbf{100.00}\\
 & Image: POPE & \textbf{76.55} / 70.50 / 87.90 & 75.85 / 75.60 / 99.50 & 76.00 / \textbf{75.80} / \textbf{99.60}\\
 & Pair: CelebA-HQ & \textbf{56.21} / 46.88 / 65.62 & 54.69 / \textbf{53.12} / \textbf{96.88} & 55.03 / \textbf{53.12} / 91.67\\
 & Pair: MMBench & \textbf{83.59} / 80.71 / 93.71 & 83.06 / 82.52 / 98.54 & 83.30 / \textbf{82.86} / \textbf{98.97}\\
 & Modality: V* & 49.74 / 39.27 / 70.16 & 43.98 / 42.93 / 97.91 & \textbf{50.79} / \textbf{50.26} / \textbf{98.95}\\
 & Modality: A-OKVQA & 77.38 / 69.69 / 81.57 & 71.83 / 71.44 / 98.86 & \textbf{79.39} / \textbf{79.13} / \textbf{99.39}\\
 & Macro-average & 66.29 / 58.93 / 80.63 & 67.08 / 66.16 / 97.80 & \textbf{69.74} / \textbf{69.20} / \textbf{98.14}\\
\midrule
\multicolumn{2}{l}{Overall macro-average} & 63.26 / 53.28 / 73.20 & 63.33 / 62.48 / 97.44 & \textbf{66.18} / \textbf{65.58} / \textbf{98.20}\\
\bottomrule\end{tabular}
\par\smallskip\begin{minipage}{0.97\textwidth}\small
Entries are AA / AC / IP (\%). AA is mean accuracy across permutations. AC counts a question as correct only if every order is correct. IP measures prediction consistency across orders. Native uses the original model. Frozen SP changes positions and masks without training. SP + KD uses the epoch-3 knowledge-distillation weights. SP is SP-V for modality exchange and SP-B elsewhere. Each setting has equal weight in the macro-averages. Bold marks the best value for each metric within a row, including ties.
\end{minipage}
\end{table*}

\subsection{Ablation Analysis of Structure and Distillation}\label{sec:ablation}

We decompose the existing results in Table~\ref{tab:recovery} into the changes associated with frozen shared-position inference, subsequent distillation, and their combined effect. Table~\ref{tab:ablation} averages the changes equally across the eight settings for each model and across all 24 model--setting comparisons. SP denotes SP-V for modality exchange and SP-B elsewhere.

\paragraph{Structure and adaptation} Frozen SP increases macro-average AC by 9.20 percentage points, with a 0.08-point change in AA. Adding KD increases AA by 2.84 points and AC by 3.10 points relative to frozen SP. The combined gains over native inference are 2.92 points in AA and 12.30 points in AC. Thus, the structural configuration primarily improves correctness sustained across input orders, while adaptation improves predictive accuracy under that configuration. 

\begin{table*}[t]\centering\small\renewcommand{\arraystretch}{1.18}
\caption{Contributions of structure and distillation.}\label{tab:ablation}
\setlength{\tabcolsep}{10pt}
\begin{tabular}{l|c|c|c}\toprule
 & Frozen SP $-$ Native & SP + KD $-$ Frozen SP & SP + KD $-$ Native\\
Model & $\Delta$AA / $\Delta$AC & $\Delta$AA / $\Delta$AC & $\Delta$AA / $\Delta$AC\\\midrule
LLaVA & -3.63 / +8.66 & +2.91 / +2.97 & -0.72 / +11.63\\
Qwen & +3.07 / +11.72 & +2.95 / +3.29 & +6.03 / +15.01\\
InternVL & +0.78 / +7.23 & +2.66 / +3.04 & +3.45 / +10.27\\
All models & +0.08 / +9.20 & +2.84 / +3.10 & +2.92 / +12.30\\
\bottomrule\end{tabular}
\par\smallskip\begin{minipage}{0.97\textwidth}\small
Entries are changes in AA / AC, in percentage points. AA is mean accuracy across permutations. AC counts a question as correct only if every order is correct. Frozen SP changes positions and masks without training. KD is knowledge distillation, using epoch-3 weights. SP is SP-V for modality exchange and SP-B elsewhere. Each model averages eight settings equally; the final row averages all 24 groups. Differences use unrounded results.
\end{minipage}\end{table*}

\paragraph{Cross-unit visibility} The modality comparisons in Figure~\ref{fig:visibility} isolate the visibility choice at fixed shared positions. SP-B lowers AA relative to native inference in all six groups, by 6.54--23.45 percentage points. Replacing blocked visibility with bidirectional cross-unit attention raises AA in five groups, as detailed in Table~\ref{tab:mask_ablation}. This pattern supports direct image--question interaction as the structural choice for modality adaptation. LLaVA on A-OKVQA provides the exception, where frozen SP-V has lower AA than SP-B.

\begin{table*}[t]\centering\small\renewcommand{\arraystretch}{1.18}
\caption{Cross-unit visibility on modality exchange.}\label{tab:mask_ablation}
\setlength{\tabcolsep}{10pt}
\begin{tabular}{l|c|c}\toprule
 & SP-B $-$ Native & SP-V $-$ SP-B\\
Model / Dataset & $\Delta$AA / $\Delta$AC & $\Delta$AA / $\Delta$AC\\\midrule
LLaVA / V* & -6.54 / +12.57 & +2.09 / +2.62\\
LLaVA / A-OKVQA & -15.07 / +2.62 & -3.28 / -3.14\\
Qwen / V* & -8.64 / +8.90 & +10.73 / +9.95\\
Qwen / A-OKVQA & -22.84 / -14.67 & +23.45 / +23.84\\
InternVL / V* & -19.11 / -9.42 & +13.35 / +13.09\\
InternVL / A-OKVQA & -23.45 / -16.42 & +17.90 / +18.17\\
\midrule
Macro-average & -15.94 / -2.74 & +10.71 / +10.75\\
\bottomrule\end{tabular}
\par\smallskip\begin{minipage}{0.97\textwidth}\small
Entries are changes in AA / AC, in percentage points, with no training. AA is mean accuracy across permutations. AC counts a question as correct only if every order is correct. SP-B blocks attention between image and question blocks. SP-V allows attention in both directions, with the same shared positions. Native uses the original model. The macro-average weights all six groups equally.
\end{minipage}\end{table*}

\subsection{Comparison with Existing Mitigation Strategies}\label{sec:comparators}

The comparison across LLaVA, Qwen, and InternVL evaluates predictive quality together with the number of input orderings required for deployment. Full-permutation majority voting (MV) \citep{zong_permutations} aggregates native predictions at inference time, whereas supervised fine-tuning (FT) \citep{positional_overload} adapts the shared-position structure through supervised training. All methods use the same eight settings and prediction protocols described in Section~\ref{sec:setup}.

\paragraph{Full-permutation majority voting} Following the permutation-aggregation strategy investigated by \citet{zong_permutations}, MV aggregates the native model's predictions over all valid permutations of each question. The candidate receiving the largest number of votes is selected, with ties resolved using a fixed, order-independent rule. For candidate-based tasks, predictions are aligned by answer content before aggregation. For CelebA-HQ, voting is performed over parsed joint attributes.

Aggregation over the complete permutation set gives an order-independent decision under the specified tie-breaking rule, so MV's AA and AC equal its voting accuracy. This decision uses 2--24 input orderings per question in the evaluated tasks.

\paragraph{Supervised fine-tuning} We compare knowledge distillation (KD) with supervised fine-tuning (FT) under the same shared-position configurations. The FT protocol is adapted from the released implementation of Positional Overload \citep{positional_overload} and applied to all three models. The implementation uses the same SEED-Bench training and validation question lists as our distillation experiments. Two configurations are trained independently on SEED-Bench MCQA: SP-B shares positions among candidate blocks and blocks cross-block attention; SP-V shares positions between the image and question blocks and permits bidirectional cross-block attention. The model trained with candidate-block SP-B is subsequently evaluated on image and image--text demonstration reordering without additional scene-specific training.

FT updates all language-model parameters, including embeddings and the output head, using supervised full-sequence cross-entropy over prompt and correct-answer text. The visual encoder and multimodal projector or merger remain frozen. Training samples input permutations, whereas KD uses fixed original orders and attention-only LoRA. Within three training epochs, validation OA selects SP-B epoch 3 for all models and SP-V epochs 3, 2, and 1 for LLaVA, Qwen, and InternVL, respectively.

We follow the authors' released MMLU fine-tuning implementation with the No Bias sampling configuration, using Adafactor at a learning rate of $5\times10^{-6}$ and an effective batch size of 64. The same training and validation lists are used for KD and FT. Warmup is adjusted to five updates for the 75-update, three-epoch schedule. Visual modules remain frozen. Table~\ref{tab:training_config} summarizes the configurations; implementation adaptations are detailed in the Supplementary Material.

\paragraph{Comparative results} Table~\ref{tab:qwen_comparators} reports macro-averages over the eight scenario--dataset settings. Attention-only KD achieves higher AA and AC than full-language-model FT on all three models. Its AA gains are 1.91, 3.61, and 3.42 percentage points on LLaVA, Qwen, and InternVL, respectively; the corresponding AC gains are 1.95, 3.69, and 4.20 points. FT achieves higher scores in some individual settings, including text A-OKVQA and demonstration-reordering MMBench on LLaVA. The Supplementary Material reports the complete per-setting comparisons.

\begin{table*}[t]
\centering\small\renewcommand{\arraystretch}{1.20}
\caption{Cross-model comparison of permutation-mitigation strategies.}\label{tab:qwen_comparators}
\setlength{\tabcolsep}{10pt}
\begin{tabular}{lcccc}\toprule
Model & Native & MV & SP + FT & SP + KD (Ours)\\\midrule
LLaVA-1.5-7B & \textbf{57.10} / 44.30 & 56.85 / \textbf{56.85} & 54.46 / 53.98 & 56.38 / 55.93\\
Qwen2.5-VL-3B & 66.38 / 56.62 & 67.23 / 67.23 & 68.79 / 67.93 & \textbf{72.41} / \textbf{71.62}\\
InternVL3.5-4B & 66.29 / 58.93 & 65.77 / 65.77 & 66.32 / 64.99 & \textbf{69.74} / \textbf{69.20}\\
\bottomrule\end{tabular}
\par\smallskip
\begin{minipage}{0.97\textwidth}\small
Entries are AA / AC (\%), averaged equally across eight settings. AA is mean accuracy across permutations. AC counts a question as correct only if every order is correct. MV is majority voting; its AA and AC equal voting accuracy. FT is supervised fine-tuning; KD is knowledge distillation. Both use SP-V for modality exchange and SP-B elsewhere. Native, FT, and KD use one input order per prediction; MV uses 2--24. FT selects the best validation OA within three epochs; KD uses epoch 3. Bold marks the best value for each metric in a row.
\end{minipage}
\end{table*}

Compared with exhaustive MV, KD increases macro-average AA and AC by 5.18 and 4.39 points on Qwen, and by 3.97 and 3.42 points on InternVL. On LLaVA, KD achieves 56.38\% AA and 55.93\% AC, compared with 56.85\% voting accuracy. KD uses one input ordering per prediction, whereas MV aggregates 2--24. These results show how structural consistency and attention-only adaptation combine predictive performance with single-order inference.

\subsection{Discussion and Limitations}\label{sec:discussion_limitations}

\paragraph{Discussion}This paper studies permutation bias in MLLMs from four parts: categorization, root-cause analysis, structural mitigation, and capability recovery. Our results lead to three main findings. First, permutation bias is not limited to candidate-answer reordering. It also occurs in image ordering, modality exchange, and image-text demonstration reordering. The metrics (OA, AA, AC, IP) provide a unified way to evaluate these scenarios. Second, the layer-wise and intervention experiments show that the bias starts at the attention output. Positions and causal masks cause structural order dependence. Finite precision can leave small differences after these structural effects are removed. Third, structural changes alone (SP-B and SP-V) can reduce permutation sensitivity, but they can lower predictive accuracy. Attention-only distillation from the frozen native model recovers this accuracy, and the final model achieves higher macro-average AC than fine-tuning on all three models and than majority voting on Qwen and InternVL.

\paragraph{Limitations}The limitations are as follows:
\noindent\textbf{1.}We evaluate three representative MLLMs (LLaVA-1.5, Qwen2.5-VL, and InternVL3.5). These models cover different architectures and show consistent patterns. Extending the same framework to more MLLMs with different visual encoders is a natural next step.
\noindent\textbf{2.}To control variables in the mechanism analysis, the diagnostic subset uses single-token candidate answers. The broader evaluation in Section 5 keeps the original candidate format and confirms the main conclusions. Applying the SP structure and distillation to more open-ended generation tasks, such as free-form VQA and captioning, is a promising direction.
\noindent\textbf{3.}The attention-only distillation requires training a student model. While it avoids the multiple forward passes of inference-time permutation aggregation, the training cost can be further optimized (e.g., by reducing the size of the training dataset or the number of trainable parameters).
\noindent\textbf{4.}Our experiments show that SP-B and SP-V achieve strong permutation invariance across the evaluated scenarios. Future work can quantify how finite-precision errors affect consistency beyond the exact-arithmetic conditions analyzed here.

\section{Conclusion}

In this paper, we study permutation bias in Multimodal Large Language Models (MLLMs) and propose a framework to mitigate it while preserving predictive accuracy. We first show that permutation bias is not limited to text-option reordering. By categorizing it into four scenarios—text-option, image, modality, and image-text demonstration reordering—we establish a unified evaluation framework with metrics (OA, AA, AC, IP) to measure prediction correctness and consistency.

Through mechanistic analysis, we find that this bias originates at the attention output. Positions and causal masks cause structural order dependence. Finite precision can leave small differences after these structural effects are removed.

Based on these findings, we propose shared positions with blocked cross-unit attention (SP-B) and extend it to bidirectional cross-unit attention (SP-V). While these structural changes reduce permutation sensitivity, they can lower predictive accuracy. To address this, we propose an attention-only knowledge distillation framework that recovers the model capability under the SP structure. Across the eight settings, our method achieves higher macro-average AC than supervised fine-tuning on all three models and than majority voting on Qwen and InternVL, while using one input ordering per prediction.

Overall, this work shows that achieving permutation-robust MLLMs requires jointly considering evaluation design, structural intervention, and capability recovery. We hope our framework provides a useful basis for building more reliable multimodal systems in future work.

\small

\section*{References}
\begingroup
\setlength{\parindent}{0pt}
\setlength{\parskip}{5pt}
\par\hangindent=1.5em\hangafter=1
Amit, G. (2026). \emph{Permutation-robust decision modeling with candidate-independent block-causal attention} {[}Preprint{]}. arXiv. \url{https://doi.org/10.48550/arXiv.2610.01601}
\par\hangindent=1.5em\hangafter=1
Atabuzzaman, Md., Alomari, H., \& Thomas, C. (2026). \emph{Investigating length bias and robustness in LVLMs for multiple-choice question answering} {[}Preprint{]}. arXiv. \url{https://doi.org/10.48550/arXiv.2610.24500}
\par\hangindent=1.5em\hangafter=1
Atabuzzaman, Md., Asgarov, A., \& Thomas, C. (2025). Benchmarking and mitigating MCQA selection bias of large vision-language models. \emph{Proceedings of the 2025 Conference on Empirical Methods in Natural Language Processing}, 33548--33562. \url{https://doi.org/10.18653/v1/2025.emnlp-main.1703}
\par\hangindent=1.5em\hangafter=1
Bai, S., Chen, K., Liu, X., Wang, J., Ge, W., Song, S., Dang, K., Wang, P., Wang, S., Tang, J., Zhong, H., Zhu, Y., Yang, M., Li, Z., Wan, J., Wang, P., Ding, W., Fu, Z., Xu, Y., \ldots{} Lin, J. (2025). \emph{Qwen2.5-VL technical report} {[}Preprint{]}. arXiv. \url{https://doi.org/10.48550/arXiv.2502.13923}
\par\hangindent=1.5em\hangafter=1
Bito, E., Ren, Y., \& He, E. (2026). One pass, any order: Position-invariant listwise reranking for LLM-based recommendation. \emph{Proceedings of the 49th International ACM SIGIR Conference on Research and Development in Information Retrieval}, 3625--3629. \url{https://doi.org/10.1145/3805712.3809952}
\par\hangindent=1.5em\hangafter=1
Cho, G., So, Y., \& Lee, J. (2026). Choices speak louder than questions. \emph{International Conference on Learning Representations}. \url{https://proceedings.iclr.cc/paper_files/paper/2026/hash/7ae2b0eedc560e3afbffd68445b8a220-Abstract-Conference.html}
\par\hangindent=1.5em\hangafter=1
Chu, X., Wang, G., Tan, Z., Chen, X., Li, Z., Mo, T., \& Li, W. (2026). Towards order fairness: Mitigating LLMs order sensitivity through dual group advantage optimization. \emph{Proceedings of the 64th Annual Meeting of the Association for Computational Linguistics (Volume 1: Long Papers)}, 4790--4805. \url{https://doi.org/10.18653/v1/2026.acl-long.219}
\par\hangindent=1.5em\hangafter=1
Egressy, B., \& Stühmer, J. (2025). \emph{Set-LLM: A permutation-invariant LLM} {[}Preprint{]}. arXiv. \url{https://doi.org/10.48550/arXiv.2505.15433}
\par\hangindent=1.5em\hangafter=1
Fatemi, M. S., Shi, Z., Mueller, K., \& Ma, T. (2026). Guided permutation scoring (GPS): Option-order permutation for multiple-choice question answering. \emph{Findings of the Association for Computational Linguistics: EMNLP 2026}, 27294--27313. \url{https://doi.org/10.18653/2026.findings-emnlp.1290}
\par\hangindent=1.5em\hangafter=1
Guan, B., Roosta, T., Passban, P., \& Rezagholizadeh, M. (2025). \emph{The order effect: Investigating prompt sensitivity to input order in LLMs} {[}Preprint{]}. arXiv. \url{https://doi.org/10.48550/arXiv.2502.04134}
\par\hangindent=1.5em\hangafter=1
Guda, B., Francis, L., Ashungafac, G. Z., Joe-Wong, C., \& Busogi, M. (2025). Quantifying and mitigating selection bias in LLMs: A transferable LoRA fine-tuning and efficient majority voting approach. \emph{Proceedings of the 14th International Joint Conference on Natural Language Processing and the 4th Conference of the Asia-Pacific Chapter of the Association for Computational Linguistics}, 2019--2038. \url{https://doi.org/10.18653/v1/2025.findings-ijcnlp.127}
\par\hangindent=1.5em\hangafter=1
Gupta, A., \& Gandelsman, Y. (2026). \emph{Test-time training for modality order consistency in vision-language models} {[}Preprint{]}. arXiv. \url{https://doi.org/10.48550/arXiv.2607.20351}
\par\hangindent=1.5em\hangafter=1
Hinton, G., Vinyals, O., \& Dean, J. (2015). \emph{Distilling the knowledge in a neural network} {[}Preprint{]}. arXiv. \url{https://doi.org/10.48550/arXiv.1503.02531}
\par\hangindent=1.5em\hangafter=1
Hu, E. J., Shen, Y., Wallis, P., Allen-Zhu, Z., Li, Y., Wang, S., Wang, L., \& Chen, W. (2022). LoRA: Low-rank adaptation of large language models. \emph{International Conference on Learning Representations}. \url{https://openreview.net/forum?id=nZeVKeeFYf9}
\par\hangindent=1.5em\hangafter=1
Karras, T., Aila, T., Laine, S., \& Lehtinen, J. (2018). Progressive growing of GANs for improved quality, stability, and variation. \emph{International Conference on Learning Representations}. \url{https://openreview.net/forum?id=Hk99zCeAb}
\par\hangindent=1.5em\hangafter=1
Kinder, L., Edman, L., Fraser, A., \& Käfer, T. (2025). Positional overload: Positional debiasing and context window extension for large language models using set encoding. \emph{Proceedings of the 63rd Annual Meeting of the Association for Computational Linguistics (Volume 1: Long Papers)}, 3896--3908. \url{https://doi.org/10.18653/v1/2025.acl-long.197}
\par\hangindent=1.5em\hangafter=1
Li, B., Ge, Y., Ge, Y., Wang, G., Wang, R., Zhang, R., \& Shan, Y. (2024). SEED-Bench: Benchmarking multimodal large language models. \emph{Proceedings of the IEEE/CVF Conference on Computer Vision and Pattern Recognition}, 13299--13308. \url{https://openaccess.thecvf.com/content/CVPR2024/html/Li_SEED-Bench_Benchmarking_Multimodal_Large_Language_Models_CVPR_2024_paper.html}
\par\hangindent=1.5em\hangafter=1
Li, R., \& Gao, Y. (2025). \emph{Anchored answers: Unravelling positional bias in GPT-2's multiple-choice questions} {[}Preprint{]}. arXiv. \url{https://doi.org/10.48550/arXiv.2405.03205}
\par\hangindent=1.5em\hangafter=1
Li, Y., Du, Y., Zhou, K., Wang, J., Zhao, W. X., \& Wen, J.-R. (2023). Evaluating object hallucination in large vision-language models. \emph{Proceedings of the 2023 Conference on Empirical Methods in Natural Language Processing}, 292--305. \url{https://doi.org/10.18653/v1/2023.emnlp-main.20}
\par\hangindent=1.5em\hangafter=1
Liu, H., Li, C., Li, Y., \& Lee, Y. J. (2024). Improved baselines with visual instruction tuning. \emph{Proceedings of the IEEE/CVF Conference on Computer Vision and Pattern Recognition}, 26296--26306. \url{https://openaccess.thecvf.com/content/CVPR2024/html/Liu_Improved_Baselines_with_Visual_Instruction_Tuning_CVPR_2024_paper.html}
\par\hangindent=1.5em\hangafter=1
Liu, Y., Duan, H., Zhang, Y., Li, B., Zhang, S., Zhao, W., Yuan, Y., Wang, J., He, C., Liu, Z., Chen, K., \& Lin, D. (2024). MMBench: Is your multi-modal model an all-around player? \emph{Computer Vision -- ECCV 2024}, 216--233. \url{https://doi.org/10.1007/978-3-031-72658-3_13}
\par\hangindent=1.5em\hangafter=1
Liusie, A., Fathullah, Y., \& Gales, M. J. F. (2024). Teacher-student training for debiasing: General permutation debiasing for large language models. \emph{Findings of the Association for Computational Linguistics: ACL 2024}, 1376--1387. \url{https://doi.org/10.18653/v1/2024.findings-acl.81}
\par\hangindent=1.5em\hangafter=1
Mishra, A., Shekhar, S., Singh, A. K., \& Chakraborty, A. (2019). OCR-VQA: Visual question answering by reading text in images. \emph{International Conference on Document Analysis and Recognition}, 947--952. \url{https://doi.org/10.1109/icdar.2019.00156}
\par\hangindent=1.5em\hangafter=1
Nowak, M., Cadet, X., \& Chin, P. (2026). ABCD: All biases come disguised. \emph{Proceedings of the 43rd International Conference on Machine Learning, Proceedings of Machine Learning Research}, \emph{306}, 94017--94047. \url{https://proceedings.mlr.press/v306/nowak26a.html}
\par\hangindent=1.5em\hangafter=1
Ok, H., Park, M., \& Lee, J. (2026). \emph{One bias score, two factors: How mitigations affect option bias in multiple-choice evaluation} {[}Preprint{]}. arXiv. \url{https://doi.org/10.48550/arXiv.2610.20993}
\par\hangindent=1.5em\hangafter=1
Pei, X., Huang, T., Ma, Y., \& Xu, C. (2025). \emph{Rethinking causal mask attention for vision-language inference} {[}Preprint{]}. arXiv. \url{https://doi.org/10.48550/arXiv.2505.18605}
\par\hangindent=1.5em\hangafter=1
Rosenthal, F., Schmidt, S., Bagdonat, T., Graf, T., Günnemann, S., \& Schwinn, L. (2026). Unexplored flaws in multiple-choice VQA make benchmarking unreliable. \emph{Findings of the Association for Computational Linguistics: EMNLP 2026}, 3412--3451. \url{https://aclanthology.org/2026.findings-emnlp.162/}
\par\hangindent=1.5em\hangafter=1
Schwenk, D., Khandelwal, A., Clark, C., Marino, K., \& Mottaghi, R. (2022). A-OKVQA: A benchmark for visual question answering using world knowledge. \emph{Computer Vision -- ECCV 2022}, 146--162. \url{https://doi.org/10.1007/978-3-031-20074-8_9}
\par\hangindent=1.5em\hangafter=1
Su, J., Ahmed, M., Lu, Y., Pan, S., Bo, W., \& Liu, Y. (2024). RoFormer: Enhanced transformer with rotary position embedding. \emph{Neurocomputing}, \emph{568}, Article 127063. \url{https://doi.org/10.1016/j.neucom.2023.127063}
\par\hangindent=1.5em\hangafter=1
Tamba, H. (2026). \emph{Position bias is hidden behind ceiling effects: A permutation diagnostic for LLM benchmarks} {[}Preprint{]}. arXiv. \url{https://doi.org/10.48550/arXiv.2607.20864}
\par\hangindent=1.5em\hangafter=1
Tan, Z., Chu, X., Li, W., \& Mo, T. (2024). \emph{Order matters: Exploring order sensitivity in multimodal large language models} {[}Preprint{]}. arXiv. \url{https://doi.org/10.48550/arXiv.2410.16983}
\par\hangindent=1.5em\hangafter=1
Wang, H., Zhao, S., Qiang, Z., Qin, B., \& Liu, T. (2024). \emph{Beyond the answers: Reviewing the rationality of multiple choice question answering for the evaluation of large language models} {[}Preprint{]}. arXiv. \url{https://doi.org/10.48550/arXiv.2402.01349}
\par\hangindent=1.5em\hangafter=1
Wang, W., Gao, Z., Gu, L., Pu, H., Cui, L., Wei, X., Liu, Z., Jing, L., Ye, S., Shao, J., Wang, Z., Chen, Z., Zhang, H., Yang, G., Wang, H., Wei, Q., Yin, J., Li, W., Cui, E., \ldots{} Luo, G. (2025). \emph{InternVL3.5: Advancing open-source multimodal models in versatility, reasoning, and efficiency} {[}Preprint{]}. arXiv. \url{https://doi.org/10.48550/arXiv.2508.18265}
\par\hangindent=1.5em\hangafter=1
Wei, S.-L., Liao, Y.-L., Huang, H.-H., \& Chen, H.-H. (2026). Do multimodal LLMs understand order? Measuring the fragility of multimodal reasoning under input order perturbations. \emph{Proceedings of the Fifteenth Language Resources and Evaluation Conference}, 9118--9128. \url{https://doi.org/10.63317/4jtpgzks8pbr}
\par\hangindent=1.5em\hangafter=1
Wu, P., \& Xie, S. (2024). V*: Guided visual search as a core mechanism in multimodal LLMs. \emph{Proceedings of the IEEE/CVF Conference on Computer Vision and Pattern Recognition}, 13084--13094. \url{https://openaccess.thecvf.com/content/CVPR2024/papers/Wu_V_Guided_Visual_Search_as_a_Core_Mechanism_in_Multimodal_CVPR_2024_paper.pdf}
\par\hangindent=1.5em\hangafter=1
Yao, J., Liu, S., Wang, Y., Mei, L., Bi, B., Ge, Y., Li, Z., \& Cheng, X. (2025). Who is in the spotlight: The hidden bias undermining multimodal retrieval-augmented generation. \emph{Proceedings of the 2025 Conference on Empirical Methods in Natural Language Processing}, 15183--15193. \url{https://doi.org/10.18653/v1/2025.emnlp-main.768}
\par\hangindent=1.5em\hangafter=1
Zhang, Q., Zhang, H., Pang, L., Zheng, H., \& Zheng, Z. (2026). Stable-RAG: Mitigating retrieval-permutation-induced hallucinations in retrieval-augmented generation. \emph{Proceedings of the 64th Annual Meeting of the Association for Computational Linguistics (Volume 1: Long Papers)}, 25907--25926. \url{https://doi.org/10.18653/v1/2026.acl-long.1188}
\par\hangindent=1.5em\hangafter=1
Zheng, J., Yuan, J., Yao, J., Gu, C., Zheng, P., \& He, G. (2026). Mitigating selection bias in large language models via permutation-aware GRPO. \emph{Proceedings of the 64th Annual Meeting of the Association for Computational Linguistics (Volume 1: Long Papers)}, 35125--35143. \url{https://doi.org/10.18653/v1/2026.acl-long.1621}
\par\hangindent=1.5em\hangafter=1
Zong, Y., Yu, T., Chavhan, R., Zhao, B., \& Hospedales, T. (2024). Fool your (vision and) language model with embarrassingly simple permutations. \emph{Proceedings of the 41st International Conference on Machine Learning}, 62892--62913. \url{https://proceedings.mlr.press/v235/zong24b.html}
\par\endgroup

\end{document}